\pdfoutput=1 
\documentclass[11pt]{article}

\usepackage[final]{acl}

\usepackage{times}
\usepackage{latexsym}
\usepackage{amssymb}

\usepackage[T1]{fontenc}

\usepackage[utf8]{inputenc}

\usepackage{microtype}

\usepackage{inconsolata}

\usepackage{graphicx}

\usepackage{multirow}                   
\usepackage{booktabs}
\usepackage{colortbl}                   
\usepackage[table,svgnames]{xcolor}     
\usepackage{amsmath}                    
\usepackage{algorithm}                  
\usepackage{algpseudocode}
\usepackage[capitalize]{cleveref}

\newcommand{\myparagraph}[1]{\vspace{8pt}\noindent{\bf #1}\hspace{2pt}}

\usepackage{marvosym}      
\usepackage[T1]{fontenc}   
\usepackage[utf8]{inputenc}
\usepackage{microtype}

\title{Watermarking for Factuality: Guiding Vision-Language Models \\Toward Truth via Tri-layer Contrastive Decoding}

\author{
 \textbf{Kyungryul Back\textsuperscript{1}},
 \textbf{Seongbeom Park\textsuperscript{1}},
 \textbf{Milim Kim\textsuperscript{1}},
 \textbf{Mincheol Kwon\textsuperscript{1}}, \\
 \textbf{SangHyeok Lee\textsuperscript{1}},
 \textbf{Hyunyoung Lee\textsuperscript{2}},
 \textbf{Junhee Cho\textsuperscript{2}},
 \textbf{Seunghyun Park\textsuperscript{3*}\textsuperscript{\Letter}},
 \textbf{Jinkyu Kim\textsuperscript{1*}\textsuperscript{\Letter}}
\\
\\
 \textsuperscript{1}CSE, Korea University \quad
 \textsuperscript{2}KT Corporation \quad
 \textsuperscript{3}Soongsil University
\\
\small{
    \{rudfuf0822, psb485, mimlmim21, kwonmc, insagur, jinkyukim\}@korea.ac.kr }
    \\
\small{
    \{lee.hyunyoung, jh.cho\}@kt.com, sh.park@ssu.ac.kr
 } \\
\small{
*Co-corresponding authors
}
}
\begin{document}

\maketitle

\begin{abstract}
Large Vision-Language Models (LVLMs) have recently shown promising results on various multimodal tasks, even achieving human-comparable performance in certain cases. Nevertheless, LVLMs remain prone to hallucinations---they often rely heavily on a single modality or memorize training data without properly grounding their outputs. To address this, we propose a training-free, tri-layer contrastive decoding with watermarking, which proceeds in three steps: (1) select a mature layer and an amateur layer among the decoding layers, (2) identify a pivot layer using a watermark-related question to assess whether the layer is visually well-grounded, and (3) apply tri-layer contrastive decoding to generate the final output. Experiments on public benchmarks such as POPE, MME and AMBER demonstrate that our method achieves state-of-the-art performance in reducing hallucinations in LVLMs and generates more visually grounded responses. Our code is available at \href{https://github.com/KR-0822/TCD}{this link}.
\end{abstract}
\section{Introduction}
\label{sec:intro}
%
\begin{figure}[tb!]
    \centering
    \includegraphics[width=\linewidth]{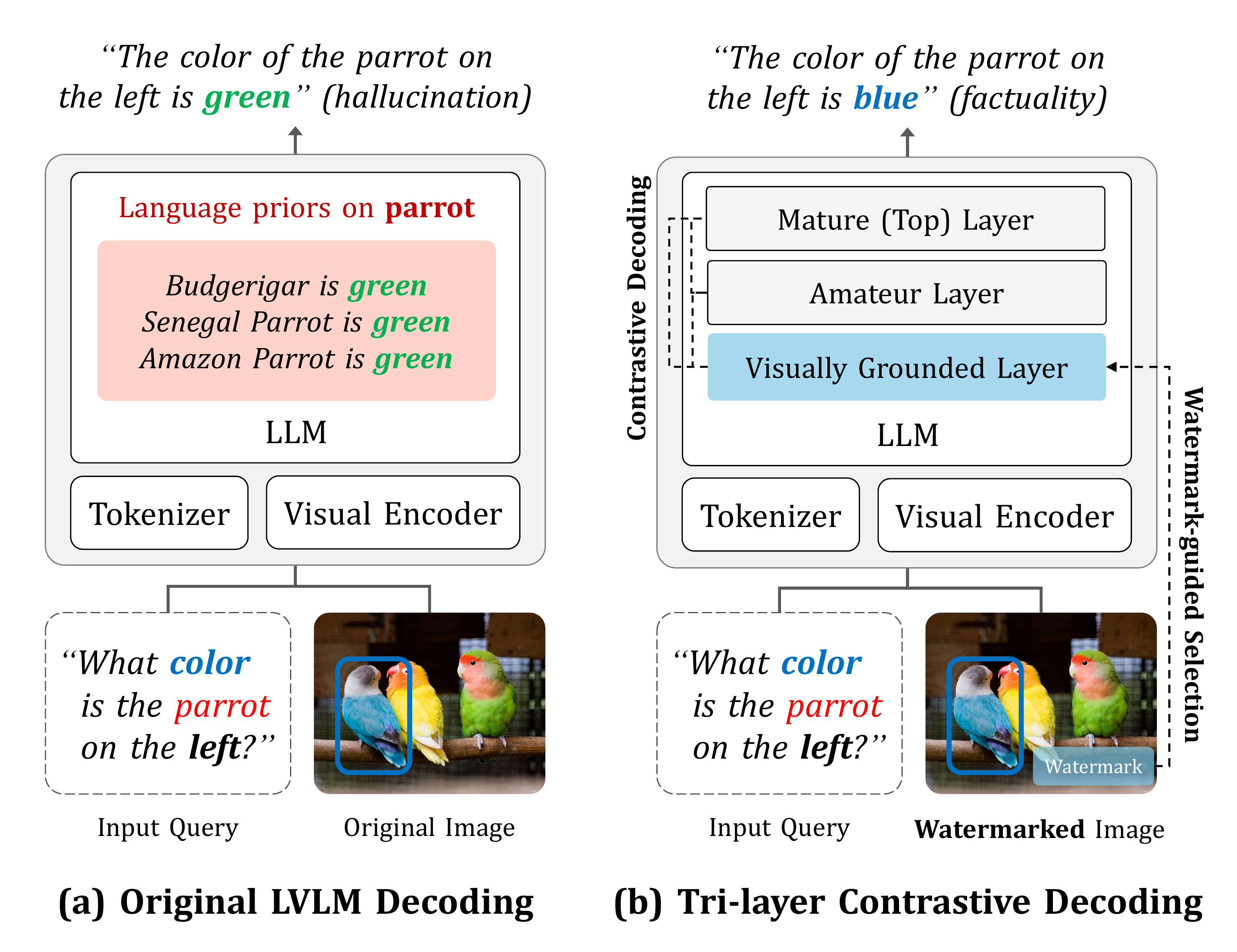}
    \vspace{-2em}
    \caption{
        Architectural comparison between (a) the conventional decoding method of LVLMs and (b) our proposed watermark-based tri-layer contrastive decoding method. To mitigate hallucinations in LVLMs, we leverage watermark for selecting visually grounded layer.
        }
    \label{fig: teaser}
    \vspace{-1em}
\end{figure}
%
Interest in Large Vision-Language Models (LVLMs) has surged recently, driven by integration of powerful large language models (LLMs) with visual encoders. This fusion enables a single model to interpret complex images and generate coherent descriptions. Recent LVLMs like LLaVA~\cite{liu2023LLaVa} and InstructBLIP~\cite{dai2023instructblip} exemplify this trend: LLaVA connects a vision encoder to an LLM via a simple projection, while InstructBLIP uses a dedicated query transformer to bridge modalities. Such LVLMs have demonstrated impressive performance on tasks including image captioning, visual question answering, and other multimodal benchmarks.

A key limitation of LVLMs is their tendency to \emph{hallucinate}---generating details absent from the image, such as naming non-existent objects or misattributing properties (see~\cref{fig: teaser}). Such hallucinations are often caused by the dominance of unimodal (language) priors. A lightweight vision module is often paired (and fine-tuned) with LLMs, which causes a modality imbalance where the language side can overwhelm the visual side~\cite{han2022visual,niu2021counterfactual,wu2022overcoming,yan2023overcoming}, outputting responses based mainly on LLMs' contextual or statistical biases. Thus, mitigating hallucinations is crucial for high-stakes applications, such as autonomous driving, medical imaging, and legal evidence analysis, where hallucinated responses could lead to severe consequences. 

To mitigate such hallucinations, various approaches have been introduced. A straightforward approach is fine-tuning or specialized training: adjusting model weights on curated datasets that emphasize image-grounded truth~\cite{gunjal2024detecting, Yin_2024, sarkar2025mitigatingobjecthallucinationmllms}, or employing Reinforcement Learning from Human Feedback (RLHF) or Direct Preference Optimization (DPO) to penalize hallucinated outputs~\cite{sun2023aligning, zhao2024hallucinationsenhancinglvlmshallucinationaware}. More recently, training-free inference-time contrastive decoding methods have emerged as efficient alternatives. For example, VCD~\cite{leng2023VCD} contrasts original and perturbed visual inputs to recalibrate the model's reliance on language priors. M3ID~\cite{Favero_2024_CVPR} boosts visual relevance via mutual information, while AVISC~\cite{woo2024dontmissforesttrees} monitors and adjusts visual attention distributions. Octopus~\cite{suo2025octopusalleviatinghallucinationdynamic} combines these strategies by dynamically selecting contrastive approaches through DPO-trained controllers. However, existing methods often overlook how visual tokens interact with language across layers, assuming final outputs suffice for grounding. To address this, we embed lightweight visual watermarks into input images and evaluate layer-wise consistency via targeted visual queries. This enables the identification of the most visually grounded intermediate layer without retraining or architectural modifications, forming the basis of our tri-layer decoding strategy.

\begin{figure*}[tb!]
    \centering
    \includegraphics[width=\linewidth]{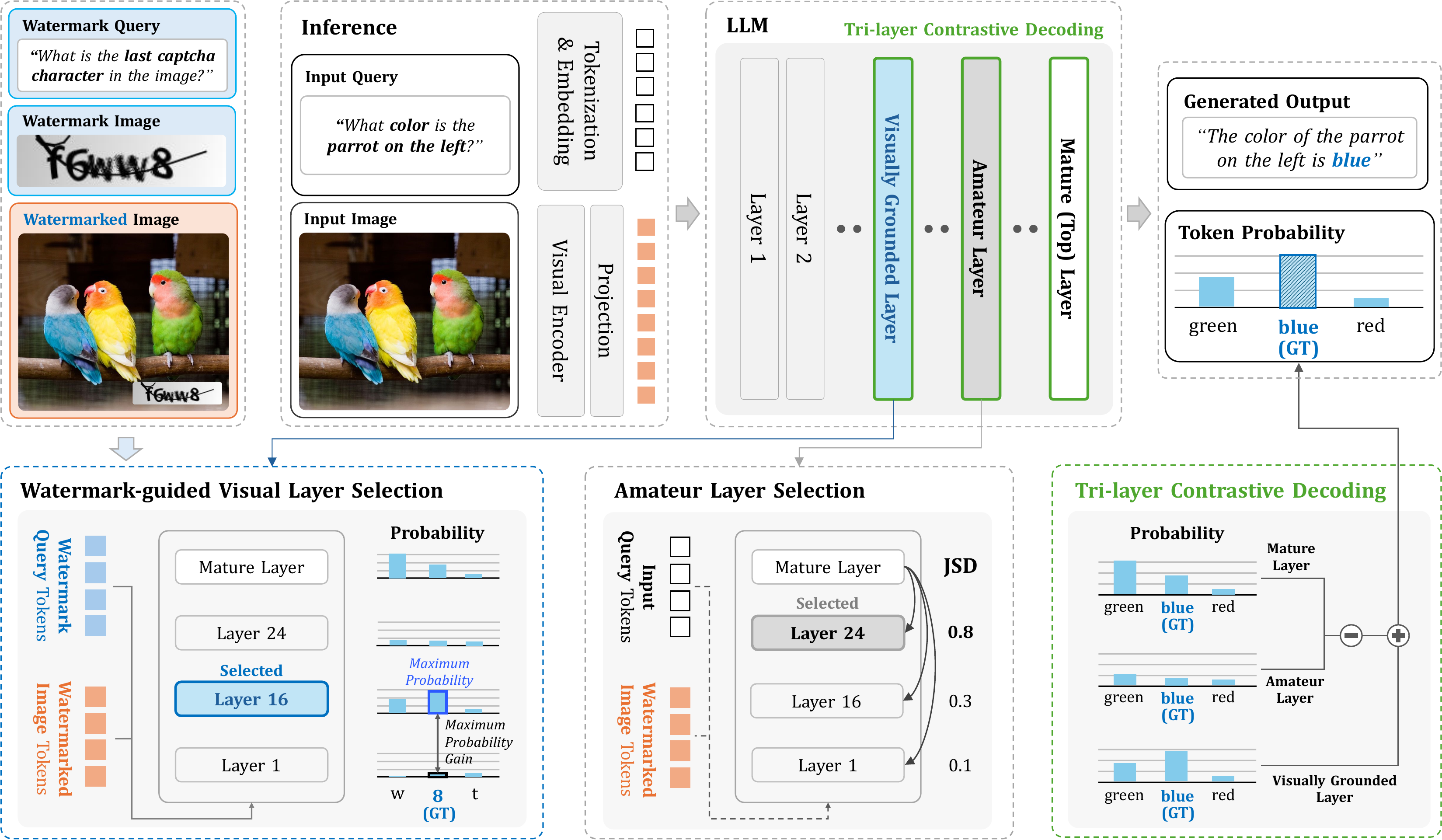}
    \caption{
        An overview of TCD, which leverages a tri-layer contrastive decoding approach by dynamically selecting and comparing following three decoding layers: (i) mature layer, (ii) amateur layer, and (iii) visually well-grounded layer. The process involves embedding a watermark into the input image, posing an ad-hoc question (e.g., ``What is the last captcha character in the image?''), and selecting the visually well-grounded layer. Note that the top layer is chosen as the mature layer, while the amateur layer is selected based on the highest JSD from the mature layer.
    }
    \label{fig: overview}
    \vspace{-.5em}
\end{figure*}

In this paper, we propose a novel training-free decoding strategy called Tri-layer Contrastive Decoding (TCD), which employs a watermark to guide the identification of the most visually grounded intermediate layer. To select this layer, we embed the watermark into the input image, query a corresponding ad-hoc question, and compare the probability distributions of an answer token across all layers. We explore \textit{maximum probability gain search}, which identifies the layer based on the probability gain of the label token prompted by the watermark between adjacent layers. Given such visually grounded layer, we decode the model using tri-layer contrastive decoding with two additional layers, i.e., mature layer defined as the top layer and an amateur layer with the maximum Jensen-Shannon Divergence (JSD) compared to the mature layer, inspired by DoLa~\cite{chuang2024dola}. We evaluate our method on widely-used hallucination benchmarks---POPE~\cite{li2023POPE}, MME~\cite{fu2024MME}, and AMBER~\cite{wang2023llm}---and show that the proposed approach achieves state-of-the-art performance across various models and settings. Detailed analyses further confirm the validity of our approach, demonstrating that watermark-guided TCD effectively mitigates hallucination. Our contributions are as follows: 
%
\begin{itemize}
    \vspace{-.25em}
    \item We propose Tri-layer Contrastive Decoding (TCD), a training-free inference framework that mitigates hallucination by contrasting three layer-wise outputs including mature, amateur, and visually grounded layer.
    \vspace{-.25em}
    \item We introduce a novel watermark-based approach to identify visually grounded layers in LVLMs by measuring visual information gain across intermediate outputs. Leveraging early-exit decoding with auxiliary visual prompts, our method enables interpretable and training-free layer selection.
    \vspace{-.25em}
    \item Extensive experiments on various benchmarks and models demonstrate the effectiveness of our proposed method, achieving state-of-the-art performance. Further analyses confirm that hallucinations are indeed alleviated, both quantitatively and qualitatively.
\end{itemize}

\section{Related Work}
\label{sec:related_work}
\vspace{-.5em}

\myparagraph{Hallucinations in LVLMs.}
%
Various large vision-language models (LVLMs) have increasingly been introduced to improve the conventional multimodal capabilities of traditional VLMs by leveraging and extending linguistic abilities of large language models (LLMs)~\cite{liu2023LLaVa, li2023blip2, bai2023qwen, yang2024qwen2}.
Despite their promising performance in various multimodal tasks, LVLMs inherit the hallucination problem that is prevalent in LLMs. Among diverse types of hallucinations, object hallucination---where the model's descriptions of objects are not well-grounded in the input image---has drawn particular attention~\cite{biten2022let,li2023POPE}.

To mitigate hallucinations in LVLMs, several approaches have been proposed. Some methods frame hallucination as a binary classification task~\cite{li2023POPE}, while others design post-hoc correction modules~\cite{zhou2023analyzing}, or apply factually augmented reinforcement learning from human feedback (RLHF)~\cite{sun2023aligning} and Direct Preference Optimization (DPO)~\cite{zhao2024hallucinationsenhancinglvlmshallucinationaware}. However, these methods typically require additional training stages and curated data.

More recently, training-free, inference-time methods have emerged to re-balance models during decoding. OPERA~\cite{Huang_2024_CVPR} penalizes over-aggregated anchor tokens in beam search. VCD~\cite{leng2023VCD} contrasts outputs from original and distorted visual inputs to reduce over-reliance on unimodal priors and statistical biases. ICD~\cite{wang-etal-2024-mitigating} suppress hallucinations by contrasting responses to perturbed instructions. M3ID~\cite{Favero_2024_CVPR} upweights image features during token sampling, and AVISC~\cite{woo2024dontmissforesttrees} reduces attention to blind tokens by monitoring visual focus. Octopus~\cite{suo2025octopusalleviatinghallucinationdynamic} dynamically selects contrastive decoding strategies using a controller trained via DPO.

All of these methods share a common philosophy: adjusting model behavior post hoc at inference time without retraining. Our proposed method aligns with this direction, but uniquely explores intermediate layers of the LVLM decoder. Instead of modifying inputs or attention distributions, we leverage the transformer's hierarchical representations to identify and utilize visually grounded layers for more reliable decoding.

\myparagraph{Layer-wise Contrastive Decoding.}
Contrastive decoding (CD) was originally introduced in LLMs to improve fluency and coherence by contrasting the outputs of a strong expert model and a weaker amateur model~\cite{li-etal-2023-contrastive}. Building on this idea, CAD~\cite{shi2024trusting} leverages surrounding context to guide generation more effectively, while ACD~\cite{gera2023benefits} enhances diversity and coherence in small LMs by fine-tuning early-layer prediction heads. Notably, DoLa~\cite{chuang2024dola} introduces a layer-wise contrastive decoding framework that dynamically selects early layers based on token complexity to reduce hallucinations.

While these studies primarily focus on LLMs, applying CD to LVLMs poses new challenges, as models must incorporate both visual and linguistic modalities. Interestingly, we observe that intermediate layers in LVLMs often generate outputs that are more visually well-grounded than those from the final decoding layer. This observation motivates our use of layer-wise contrastive decoding as a potential solution for mitigating hallucinations. 

However, identifying visually grounded layers in a training-free setting remains difficult. To address this, we propose leveraging watermarks---perturbations embedded into the input image that do not alter the final output but serve as cues for judging whether an intermediate layer is visually grounded.
\section{Method}

Given a visual context $v$ (e.g., an image) and a textual query $x$, LVLMs generate a textual response $y$.
The response $y=\{y_1, y_2,~\dots, y_{T}\}$ is calculated in an auto-regressive manner, where each token is predicted sequentially based on the preceding tokens, and $T$ represents the total number of tokens in the generated response.
Formally, the token probability distribution at each time step $t\in[1,T]$ can be formulated as follows:
%
\begin{equation}
\small
\label{eq: next-token prediction}
    p_{\theta}(y_t \mid x, v, y_{1:t-1}) = \frac{\text{exp}(\text{z}_{\theta}(y_{t} \mid x, v, y_{1:t-1})/\tau)}{\Sigma_{y'_{t}\in \mathcal{Y}}\text{exp}(\text{z}_{\theta}(y'_{t} \mid x, v, y_{1:t-1})/\tau)},
\end{equation}
where $\theta$ denotes model parameters, $z$ represents the logit of a layer, $\tau$ is a temperature for logit scaling, and $y'_{t}$ is a token in vocabulary set $\mathcal{Y}$.
Output token selection, or decoding, determines the final generated response $y$ by selecting tokens from the probability distribution in~\cref{eq: next-token prediction}.
Common decoding strategies include greedy decoding~\cite{sutskever2014sequence}, beam search~\cite{bahdanau2014neural}, and top-$k$ sampling~\cite{fan2018hierarchical}.


Despite the effectiveness of these decoding strategies, a critical challenge remains: \textit{hallucination}.
In the context of LVLMs, even if the probability distribution $p_{\theta}$ assigns a high likelihood, a token $y_t$ is considered hallucinated if it lacks sufficient grounding in the provided textual query $x$ or visual context $v$.
To this end, we propose a novel tri-layer contrastive decoding with a watermark-guided visual layer selection scheme.
This approach aims to realign the model's token probability distribution with the factual constraints in $x$ and $v$, thereby reducing the incidence of hallucinations in the generated output.
An overview of our proposed method is shown in~\cref{fig: overview}.

\subsection{Watermark-Guided Layer Selection}

To mitigate hallucinations in LVLMs, we first select the most visually representative layer through watermark-based verification.
The key intuition is that the visual information in LVLMs evolves across layers, which aligns with observations from prior work on LLMs~\cite{chuang2024dola}.

\myparagraph{Watermark Integration.}
To identify a visually informative layer, a novel question emerges: \emph{how can we identify a layer as visually informative, while preserving the visual representations of an input image?}
This motivates us to design a watermark-based verification approach that can be seamlessly integrated with an input image and simultaneously provides a cue about the information in each layer.
Specifically, we embed a watermark image into the input image and prepend a watermark question to the textual query.
The watermark serves to examine each layer’s representation in the model by leveraging image data related to vision-language tasks, such as CAPTCHAs.
Formally, given a watermark image $\mathcal{I}_{\text{wm}}$ and a watermark textual query $x_{\text{wm}}$, the visual context $v$ and the textual query $x$ are generated as follows:
%
\begin{gather}
\label{eq: watermark image}
    v = f_{\text{visual}}(\mathcal{I}_{\text{org}} + \alpha \mathcal{I}_{\text{wm}})
\end{gather}
where $f_{\text{visual}}$ is a visual encoder, $\mathcal{I}_{\text{org}}$ is the input image, $x_{\text{org}}$ is the input text query, and $\alpha$ is the opacity hyperparameter for the watermark.
For clarity, we construct a watermark question that has a fixed length and a clear answer (e.g., \textit{``What is the last number in the CAPTCHA image?''}).
In this section, we assume that $\mathcal{I}_{\text{wm}}$ is appropriately preprocessed (e.g., in terms of size and position) for the integration.
For further details and analyses of watermark preprocessing, please see~\cref{subsec: experimental setup}, as well as~\cref{alg: watermark} and~\cref{fig: CAPTCHA}, both located in the Appendix.
%

\begin{table*}[t]
    \centering
    \setlength{\tabcolsep}{15pt}
    \renewcommand{\arraystretch}{0.8}
    \resizebox{\linewidth}{!}{%
    \begin{tabular}{@{}llcccccc@{}}
        \toprule
        \multirow{2}{*}{\textbf{LVLM}} & \multirow{2}{*}{\textbf{Method}} & \multicolumn{2}{c}{\textbf{MSCOCO}} & \multicolumn{2}{c}{\textbf{OKVQA}} & \multicolumn{2}{c}{\textbf{GQA}} \\
        \cmidrule(lr){3-4} \cmidrule(lr){5-6} \cmidrule(l){7-8}
        & & Acc.($\uparrow$) & F1($\uparrow$) & Acc.($\uparrow$) & F1($\uparrow$) & Acc.($\uparrow$) & F1($\uparrow$) \\
        \midrule
        \multicolumn{8}{c}{\textit{Referenced Results (Not Directly Comparable)}} \\
        \midrule
        \multirow{3}{*}{LLaVA-v1.5}
            & EOS & 86.80 & 86.00  & -    & -    & -    & -    \\
            & HA-DPO & 86.63 & 86.87  & -    & -    & -    & -    \\
            & Octopus & 85.79 & 83.44 & -    & -    & -    & -    \\
        \midrule
        \multirow{3}{*}{InstructBLIP}
            & OPERA & 79.13 & 79.74 & -    & -    & -    & -    \\
            & HA-DPO $^*$ & 85.43 & 85.64 & -    & -    & -    & -    \\
            & Octopus & 84.79 & 83.43 & -    & -    & -    & -    \\
        \midrule
        \multicolumn{8}{c}{\textit{Comparable Results (Training-Free Contrastive Decoding)}} \\
        \midrule
        \multirow{6}{*}{LLaVA-v1.5}
            & Base       & 82.04 & 80.42 & 75.58 & 79.23 & 74.39 & 78.58 \\
            & + ICD      & 83.26 & 82.53 & -    & -    & -    & -    \\
            & + VCD      & 82.96 & 81.81 & 74.72 & 78.87 & 74.10 & 78.70 \\
            & + M3ID     & 82.57 & 80.26 & 76.16 & 79.91 & 74.60 & 78.99 \\
            & + AVISC    & \underline{83.39} & \underline{81.01} & \underline{77.47} & \underline{80.87} & \underline{76.33} & \underline{80.40} \\
            \rowcolor{gray!20} \cellcolor{white} & + TCD (Ours) & \textbf{87.00} & \textbf{86.65} & \textbf{86.46} & \textbf{87.07} & \textbf{85.47} & \textbf{85.44} \\
        \midrule
        \multirow{6}{*}{InstructBLIP}
            & Base       & 79.14 & 79.31 & 74.93 & 77.86 & 73.84 & 76.70 \\
            & + ICD      & 79.14 & 79.92 & -    & -    & -    & -    \\
            & + VCD      & 79.46 & 79.49 & 75.59 & 78.28 & 75.36 & 77.87 \\
            & + M3ID     & 80.59 & 80.15 & 75.83 & 78.80 & 74.68 & 77.62 \\
            & + AVISC    & \underline{84.04} & \underline{82.62} & \underline{80.92} & \underline{82.62} & \underline{79.85} & \underline{80.98} \\
            \rowcolor{gray!20} \cellcolor{white} & + TCD (Ours) & \textbf{84.10} & \textbf{83.88} & \textbf{82.88} & \textbf{84.33} & \textbf{80.96} & \textbf{82.39} \\ 
        \bottomrule
    \end{tabular}}
    \vspace{-.5em}
    \caption{
        Performance comparison on discriminative tasks (ALL split) across the POPE-MSCOCO, A-OKVQA, and GQA datasets. The best results are shown in \textbf{bold} and the second-best is \underline{underlined}. $^*$ Denotes InstructBLIP with the Vicuna-13B backbone; all other models are based on Vicuna-7B. Complete results for the Random, Popular, and Adversarial subsets are provided in Appendix~\cref{tab: app_pope_mscoco,tab: app_pope_okvqa,tab: app_pope_gqa}.
    }
    \label{tab: main_pope}
    \vspace{-1em}
\end{table*}

\myparagraph{Layer Selection in LVLMs.}
Our goal is to identify the decoding layer $l_v$ that contains visually informative representations using the watermark-integrated inputs $x$ and $v$.
We select a layer based on the probability distribution $p_{\theta}$ in~\cref{eq: next-token prediction}, where the logit $z$ is computed using the hidden representation $h_{t-1}$ and the vocabulary head $g$, i.e., $z = g(h_{t-1})$.
Although $z$ is often computed using the last layer representation for final output generation (i.e., $z = g(h_{t-1}^{(L)})$), it is also possible to apply the language head $g$ to intermediate layers---an approach known as early exit~\cite{teerapittayanon2016branchynet,schuster2022confident,chuang2024dola}---to leverage a model's implicit factual knowledge.

Given the watermark-integrated textual query $x$ and visual context $v$, the hidden representation of layer $l$, $h_{t-1}^{(l)}$, is generated by first processing the input through the embedding layer $f_{\text{embed}}$ and then through a series of transformer layers $f_{\text{trans}}^{(l)}$:
%
\begin{gather}
\label{eq: hidden representation}
    h_{t-1}^{(0)} = f_{\text{embed}}(x, v, y_{1:t-1}, )\\
    h_{t-1}^{(l)} = f_{\text{trans}}^{(l)}(h_{t-1}^{{(l-1)}}),~~l \in \{1,\,2,\,\dots,\,L\},
\end{gather}
where $L$ is the total number of transformer layers.
Using these hidden representations, we compute the layer-wise token probability distribution $p_{\theta}^{(l)}$:
%
\begin{equation}
\label{eq: layer-wise prob}
     p_{\theta}^{(l)}=\mathrm{softmax}(z_{\theta}^{(l)})=\mathrm{softmax}(g(h_{t-1}^{(l)})).
\end{equation}

\myparagraph{Watermark-Guided Visual Layer Selection.}
Given the layer-wise probability distribution of the watermark label $y_{\text{wm}}$, we identify the layer with the greatest probability increase compared to the previous layer---referred to as \emph{maximum probability gain search}---as formulated as follows:
%
\begin{equation}
\label{eq: visual layer selection}
    l_v =        \operatorname{argmax}_{l}\; \Delta p_{\theta}^{(l)}(y_{\text{wm}} \mid x, v) 
\end{equation}
where $\Delta$ denotes the difference in probability between adjacent layers:
%
\begin{equation}
\label{eq:delta}
\Delta p_{\theta}^{(l)} =
\begin{cases}
p_{\theta}^{(l)} - p_{\theta}^{(l-1)}, & \text{(i)} \\[1ex]
\log\left(\frac{p_{\theta}^{(l)}}{p_{\theta}^{(l-1)}}\right). & \text{(ii)}
\end{cases}
\end{equation}
%
%
%
%
therefore, $p_{\theta}^{(l)}$ is measured using the first sequence of generated tokens (for simplicity, we ignore the special tokens).

\subsection{Tri-layer Contrastive Decoding}

In our framework, we leverage the visual layer $l_v$ as a reference probability distribution for contrastive decoding.
Following prior work~\cite{chuang2024dola}, we define the final layer $L$ as a mature layer and use it as an anchor distribution.
The negative distribution, $l_a$ (referred to as an amateur layer), is selected based on the highest Jensen-Shannon Divergence (JSD) between the distributions of the intermediate layers and the anchor distribution:
%
\begin{equation}
\label{eq: JSD}
    l_a = \operatorname{argmax}_{l} \mathrm{JSD}\bigl(p_{\theta}^{(L)},\,p_{\theta}^{(l)}\bigr),
\end{equation}
where $l \in \{1,\,2,\,\dots,\,L-1\}$ is an intermediate layer index.
Note that a high JSD implies that such a layer offers an alternative perspective prior to the final layer's information accumulation, making it a strong candidate for contrastive decoding.

\myparagraph{Constraints on Contrastive Decoding.}
When a token exhibits high confidence in both the mature layer $L$ and the amateur layer $l_a$, the contrastive decoding process may reduce the relative difference between probabilities, making a previously certain decision ambiguous.
To address this, we adopt the Adaptive Plausibility Constraint (APC), following prior works~\cite{li-etal-2023-contrastive, leng2023VCD, chuang2024dola}. Formally, we define the set of viable tokens $\mathcal{V}$ as follows:
%
\begin{equation}
\small
\label{eq: APC} 
    \mathcal{V}(y_t \mid y_{1:t-1}) = \left\{ y_t \in \mathcal{Y} \mid p_{\theta}^{(L)}(y_t) \geq \beta \max\limits_{y'_t} p_{\theta}^{(L)}(y'_t) \right\}
\end{equation} 
where $\beta \in [0,1]$ is a hyperparameter that determines the threshold for plausible token selection. 

\myparagraph{Final Output Generation.}
To generate the final response $y$, we first define a constraint function $F(\cdot)$ to leverage APC on the input tokens:
\begin{equation}
    \small
    \label{eq: APC_function}
        F(z_{\theta}(y_t)) =
        \begin{cases}
        z^{(L)} - z^{(l_a)} + \lambda z^{(l_v)} & \text{if } y_t \in \mathcal{V}(y_t \mid y_{<t}) \\
        -\infty & \text{otherwise}.
        \end{cases}
\end{equation}
This formulation ensures that contrastive decoding effectively integrates visual grounding while avoiding false positives (implausible tokens receiving disproportionately high scores) and false negatives (valid tokens being overlooked due to contrastive decoding effects) through the application of APC, thereby reducing hallucinations in generated responses.
Finally, we generate the token sequence $y$ using the refined logits under the APC constraint:
%
\begin{equation}
\label{eq: final}
y \sim \hat{p}_{\theta} = \mathrm{argmax}(F(z_{\theta}(y_t))).
\end{equation}


%

\section{Experiments}
 \begin{table}[t]
    \begin{center}
    \setlength{\tabcolsep}{5pt}
    \renewcommand{\arraystretch}{0.9}
    \resizebox{\linewidth}{!}{%
         \begin{tabular}{@{}llcccc|c@{}}
         \toprule
         \multirow{2}{*}{LVLM} & \multirow{2}{*}{Method} & \multicolumn{2}{c}{Object-level} & \multicolumn{2}{c|}{Attribute-level} & \multirow{2}{*}{Total($\uparrow$)} \\
          \cmidrule(lr){3-4} \cmidrule(lr){5-6}
         & & Existence($\uparrow$) & Count($\uparrow$) & Position($\uparrow$) & Color($\uparrow$) & \\
         \midrule
         \multirow{5}{*}{LLaVA-v1.5} 
            &  Base    & 173.57 & 110.00 & 100.47 & 125.24 & 509.28 \\
            & + VCD    & 172.14 & \underline{117.14} & 103.33 & 119.52 & 512.14 \\
            & + M3ID   & 178.33 & 107.22 & 96.39  & 127.50 & 509.44 \\
            & + AVISC  & \textbf{189.29} & 104.76 & \underline{106.19} & \underline{127.86} & \underline{528.09} \\
            \rowcolor{gray!20} \cellcolor{white}
            & + TCD (Ours) & \underline{185.00} & \textbf{158.3} & \textbf{135.0} & \textbf{175.0} & \textbf{653.30} \\
         \midrule
         \multirow{5}{*}{InstructBLIP}
            &   Base   & 170.19 & 89.52 & 67.62 & 114.76 & 442.09 \\
            & + VCD    & 172.62 & \underline{98.33} & 71.90 & 117.14 & 459.99 \\
            & + M3ID   & 173.89 & 89.72 & 72.72 & 110.56 & 446.88 \\
            & + AVISC  & \textbf{184.76} & 82.85 & \underline{74.76} & \underline{131.43} & \underline{473.80} \\
            \rowcolor{gray!20} \cellcolor{white}
            & + TCD (Ours)  & \underline{180.00} & \textbf{116.67} & \textbf{76.66} & \textbf{158.33} & \textbf{531.67} \\
         \bottomrule
         \end{tabular}
    }
    \vspace{-.5em}
    \caption{
        Performance comparison on the discriminative task using the coarse-grained perception subset of the MME~\cite{fu2024MME} benchmark.
       }
    \label{tab: main_mme_table}
    \end{center}
    \vspace{-1em}
 \end{table}

\subsection{Experimental Setup} 
\label{subsec: experimental setup}
\vspace{-.5em}
\myparagraph{Benchmarks and LVLMs.}
To evaluate LVLM's hallucination performance, we use three widely used benchmarks: POPE~\cite{li2023POPE}, a perception subset of MME~\cite{fu2024MME}, and AMBER~\cite{wang2023llm}. Following previous works~\cite{leng2023VCD, woo2024dontmissforesttrees, suo2025octopusalleviatinghallucinationdynamic}, we evaluate the discriminative task on POPE, MME and generative task on AMBER.
\textbf{POPE} is used to assess object hallucination by querying whether a specific object exists in an image, using a balanced set of positive and negative queries.
It employs three sampling strategies---adversarial, popular, and random---across three datasets (i.e., MS-COCO~\cite{lin2014COCO}, A-OKVQA~\cite{schwenk2022okvqa}, and GQA~\cite{hudson2019gqa}), thereby generating a total of 27,000 query-answer pairs.
In addition, we use the \textbf{MME} benchmark to evaluate LVLMs on perception-related tasks. Following prior work~\cite{yin2024woodpecker, leng2023VCD}, we focus on object-level hallucination (existence and count) and attribute-level hallucination (position and color). For generative tasks, we utilize \textbf{AMBER}, an automated LLM-free multi-dimensional benchmark. Four metrics including Cover, Hal, Cog, and CHAIR~\cite{rohrbach-etal-2018-object} are used to measure the generation quality of our method. Specifically, AMBER compares generated object mentions against human-annotated ground truth to evaluate object coverage (Cover), hallucination frequency (Hal), cognitively plausible hallucinations (Cog), and the proportion of hallucinated objects (CHAIR), providing a comprehensive and cost-efficient assessment of hallucination.
In our experiments, we evaluate our method on two widely used LVLMs, LLaVA-1.5~\cite{liu2023LLaVa} and InstructBLIP~\cite{dai2023instructblip}, both using Vicuna-7B as the backbone. We also apply our method to generative tasks using DeepSeek-VL2\cite{wu2024deepseekvl2mixtureofexpertsvisionlanguagemodels}, a model with a Mixture of Expert (MoE) architecture, thereby demonstrating the robustness of TCD on a stronger backbone.

\begin{table}[t]
    \begin{center}
    \setlength{\tabcolsep}{2pt}
    \renewcommand{\arraystretch}{0.9}
    \resizebox{\linewidth}{!}{%
    \begin{tabular}{@{}llcccc@{}}
        \toprule
        LVLM & Method & CHAIR($\downarrow$) & Cover.($\uparrow$) & HalRate($\downarrow$) & Cog.($\downarrow$) \\
        \midrule
        \multicolumn{6}{c}{\textit{Referenced Results (Not Directly Comparable)}} \\
        \midrule
        \multirow{4}{*}{LLaVA-v1.5}
                         & EOS     & 5.1 & 49.1 & 22.7 & 2.0 \\
                         & HA-DPO  & 6.7 & 49.8 & 30.9 & 3.3 \\
                         & HALVA   & 6.6 & 53.0 & 32.2 & 3.4 \\
                         & Octopus & 4.8 & 49.2 & 23.4 & 1.2 \\
        \midrule
        \multicolumn{6}{c}{\textit{Comparable Results (Training free Contrastive Decoding)}} \\
        \midrule
        \multirow{5}{*}{LLaVA-v1.5}
                         & Base           & 8.0 & 44.5 & 31.0 & 2.2 \\
                         & + VCD      & 6.7 & 46.5 & 27.8 & 2.0 \\
                         & + M3ID     & \underline{6.0} & \textbf{48.9} & 26.0 & \textbf{1.5} \\
                         & + AVISC    & 6.3 & 46.6 & \underline{25.6} & 2.0 \\
                         \rowcolor{gray!20} \cellcolor{white} & + TCD (Ours) & \textbf{4.4} & \underline{47.2} & \textbf{19.2} & \underline{1.7} \\
                         
        \cmidrule(lr){1-6}
        \multirow{5}{*}{InstructBLIP}
                         & Base           & 8.4 & 46.4 & 31.1 & 2.6 \\
                         & + VCD      & 7.6 & \underline{47.7} & 29.9 & \textbf{2.2} \\
                         & + M3ID     & 6.9 & 47.2 & \underline{27.5} & \textbf{2.2} \\
                         & + AVISC    & \underline{6.7} & 46.7 & 28.0 & 2.6 \\
                        \rowcolor{gray!20} \cellcolor{white} & + TCD (Ours) & \textbf{6.3} & \textbf{48.8} & \textbf{26.8} & \underline{2.3} \\
        \midrule
        \multicolumn{6}{c}{\textit{Appliance to a Stronger Backbone}} \\
        \midrule
        \multirow{3}{*}{DeepSeek-VL2-Tiny~~}
                         & Base           & \underline{3.8} & \underline{56.8} & \underline{18.2} & \underline{1.0} \\
                         & + VCD$^*$           & 4.7 & \textbf{56.9} & 22.4 & 1.3 \\
                         \rowcolor{gray!20} \cellcolor{white} & + TCD (Ours) & \textbf{3.6} & 56.3 & \textbf{16.5} & \textbf{0.8} \\
        \bottomrule
        \end{tabular}
    }
    \end{center}
    \vspace{-.5em}
    \caption{
        Performance comparison on the generative task using the AMBER~\cite{wang2023llm} benchmark. $^*$ Indicates results implemented using the official code.
    }
    \label{tab: main-amber}

    \vspace{-1em}
\end{table}

\myparagraph{Implementation Details.}
Following prior work~\cite{chuang2024dola,leng2023VCD}, we set $\beta=0.1$ for stable CD and use 20 candidate layers for both LVLMs, except in the case of MME evaluation for InstructBLIP. Other parameters such as $\lambda $ and question templates, are provided in the~\cref{app: C}. We leverage simple yet effective CAPTCHA~\cite{captcha} dataset for watermark verification. Further, to seamlessly integrate a watermark into the input image, we apply light preprocessing (e.g., position, size, and opacity). The watermark is placed in the bottom-right corner with opacity $\alpha=0.8$. Additional implementation details are provided in the \cref{watermark implementation} and~\cref{fig: CAPTCHA}.


\subsection{Experimental Results}
\label{subsec: quant_analysis}

\begin{table}[tb!]
    \setlength{\tabcolsep}{0.2\tabcolsep}
    \scriptsize
    \centering
    \begin{tabular*}{\linewidth}{@{\extracolsep{\fill}}lllcc}
        \toprule
        Model & Setting & Decoding & Accuracy($\uparrow$) & F1($\uparrow$) \\
        \midrule
        \multirow{12}{*}{LLaVA1.5 (7B)} 
        & \multirow{4}{*}{Random} 
        & Greedy     & 85.87 & 84.37 \\
        &            & + AL         & 87.70 \textcolor{red}{(+1.83)} & 86.37 \textcolor{red}{(+2.00)} \\
        &            & + VL         & 89.90 \textcolor{red}{(+4.03)} & 89.48 \textcolor{red}{(+5.11)} \\
        &            & + AL+VL~     & 89.50 \textcolor{red}{(+3.63)} & 88.89 \textcolor{red}{(+4.52)} \\
        \cmidrule{2-5}
        & \multirow{4}{*}{Popular} 
        & Greedy     & 84.10 & 82.75 \\
        &            & + AL         & 86.63 \textcolor{red}{(+2.53)} & 85.34 \textcolor{red}{(+2.59)} \\
        &            & + VL         & 87.73 \textcolor{red}{(+3.63)} & 87.51 \textcolor{red}{(+4.76)} \\
        &            & + AL+VL~     & 87.60 \textcolor{red}{(+3.50)} & 87.14 \textcolor{red}{(+4.39)} \\
        \cmidrule{2-5}
        & \multirow{4}{*}{Adversarial} 
        & Greedy     & 81.03 & 80.10 \\
        &            & + AL         & 84.27 \textcolor{red}{(+3.24)} & 83.18 \textcolor{red}{(+3.08)} \\
        &            & + VL         & 83.63 \textcolor{red}{(+2.60)} & 84.00 \textcolor{red}{(+3.90)} \\
        &            & + AL+VL~     & 83.90 \textcolor{red}{(+2.87)} & 83.92 \textcolor{red}{(+3.82)} \\
        \midrule
        \multirow{12}{*}{LLaVA1.5 (13B)} 
        & \multirow{4}{*}{Random} 
        & Greedy     & 85.47 & 84.32 \\
        &            & + AL         & 87.03 \textcolor{red}{(+1.56)} & 85.84 \textcolor{red}{(+1.52)} \\
        &            & + VL         & 90.37 \textcolor{red}{(+4.90)} & 90.52 \textcolor{red}{(+6.20)} \\
        &            & + AL+VL~     & 90.23 \textcolor{red}{(+4.76)} & 89.20 \textcolor{red}{(+4.88)} \\
        \cmidrule{2-5}
        & \multirow{4}{*}{Popular} 
        & Greedy     & 84.07 & 82.89 \\
        &            & + AL         & 87.03 \textcolor{red}{(+2.96)} & 85.84 \textcolor{red}{(+2.95)} \\
        &            & + VL         & 88.43 \textcolor{red}{(+4.36)} & 88.82 \textcolor{red}{(+5.93)} \\
        &            & + AL+VL~     & 89.70 \textcolor{red}{(+5.63)} & 89.20 \textcolor{red}{(+6.31)} \\
        \cmidrule{2-5}
        & \multirow{4}{*}{Adversarial} 
        & Greedy     & 81.90 & 81.14 \\
        &            & + AL         & 85.07 \textcolor{red}{(+3.17)} & 84.03 \textcolor{red}{(+2.89)} \\
        &            & + VL         & 82.70 \textcolor{red}{(+0.80)} & 84.16 \textcolor{red}{(+3.02)} \\
        &            & + AL+VL~     & 85.87 \textcolor{red}{(+3.97)} & 85.79 \textcolor{red}{(+4.65)} \\
        \bottomrule
    \end{tabular*}
    \vspace{-.5em}
    \caption{
        Effect of the components of the proposed contrastive decoding method: amateur layer (AL) and watermark-based visual layer (VL). We use the LLaVa-1.5 backbone on the POPE-MSCOCO benchmark. Performance gains are highlighted in \textcolor{red}{red}.
        } 
    \label{tab: main_13b}
    \vspace{-1em}
\end{table}

\myparagraph{Comparison with SOTA Approaches.} 
To validate the effectiveness of our method, we conduct evaluations using various benchmarks, models, and decoding methods. 
We use instruction fine-tuned LVLMs (referred to as ``Base'' in the tables), along with ICD, VCD, M3ID and AVISC, as our training-free contrastive decoding baselines.
We additionally compare against EOS~\cite{yue-etal-2024-less}, HA-DPO~\cite{zhao2024hallucinationsenhancinglvlmshallucinationaware}, HALVA~\cite{sarkar2025mitigating}, and Octopus, which require additional training or external models, and serve as reference methods.

As shown in~\cref{tab: main_pope}, TCD clearly outperforms the baselines and achieves state-of-the-art performance across all three subsets of POPE~\cite{li2023POPE}, in terms of both accuracy and F1 score. While Octopus combines all three baseline methods and requires additional DPO training, TCD still surpasses it---achieving higher performance for the LLaVA model and in F1 score for InstructBLIP.

The efficacy of our method in mitigating hallucinations is further confirmed in \cref{tab: main_mme_table}, while outperforming the baselines in object and attribute level. We provide full perception task score in the Appendix~\cref{tab:sub_mme}. For generative task, our method successfully mitigated hallucinations lowering the CHAIR score and Hallucination Rate of AMBER bench with huge margin as shown in \cref{tab: main-amber}. We also show our method's scalability using latest LVLM, DeepSeek-VL-Tiny \cite{wu2024deepseekvl2mixtureofexpertsvisionlanguagemodels} with stronger backbone for both visual encoder and LLM compared with LLaVA-v1.5 and InstructBLIP. 
%



\begin{figure}[tb!]
    \centering
    \includegraphics[width=\linewidth]{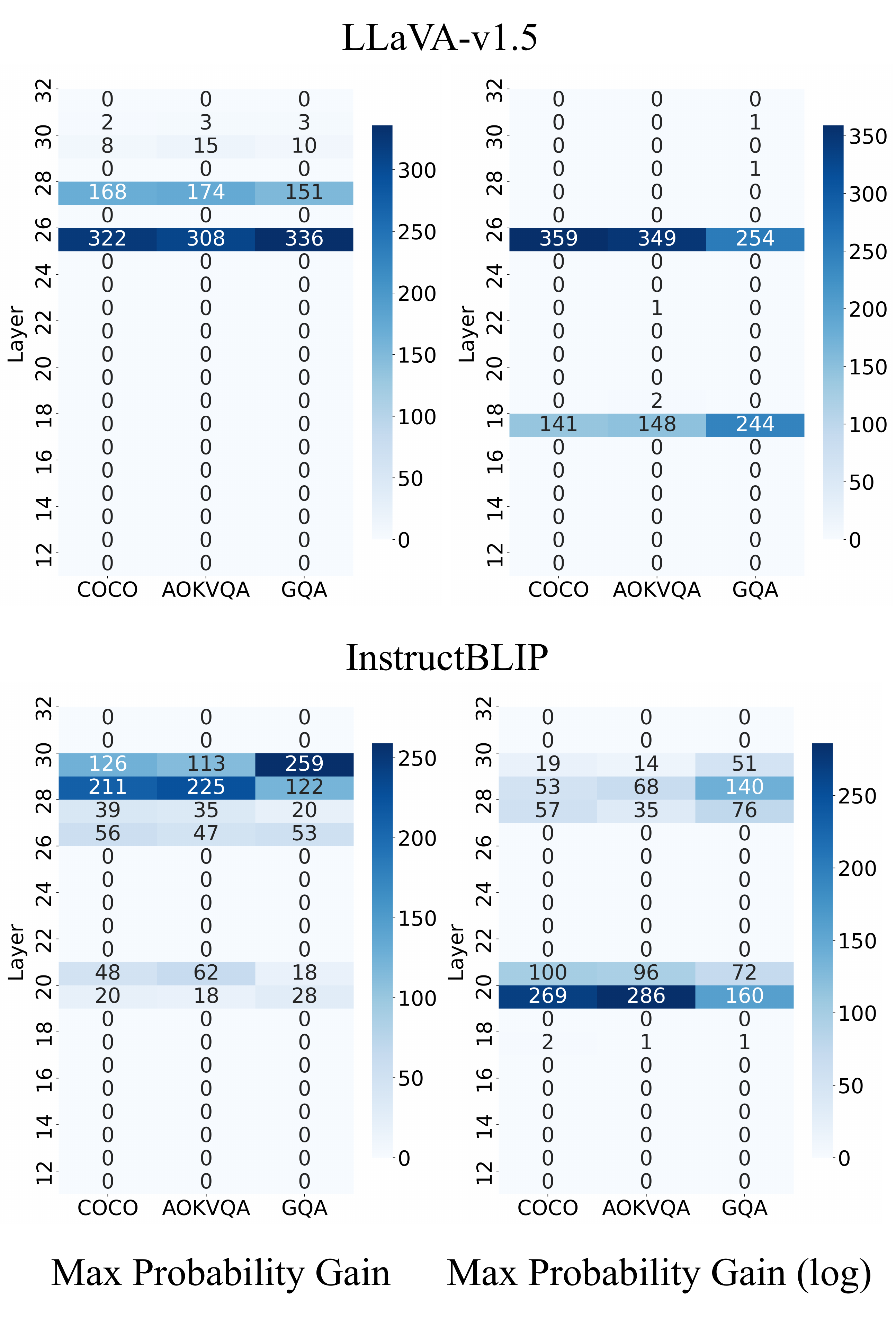}
    \vspace{-1em}
    \caption{
        Heatmaps of each LVLMs' layers selected with POPE datasets. (COCO, AOKVQA and GQA). As shown, the proposed methods showed consistent layer selection with low variance.
    }
    \vspace{-.75em}
    \label{fig: LLaVA_Layer}
\end{figure}

\begin{figure*}[tb!]
    \centering
    \includegraphics[width=.95\linewidth]{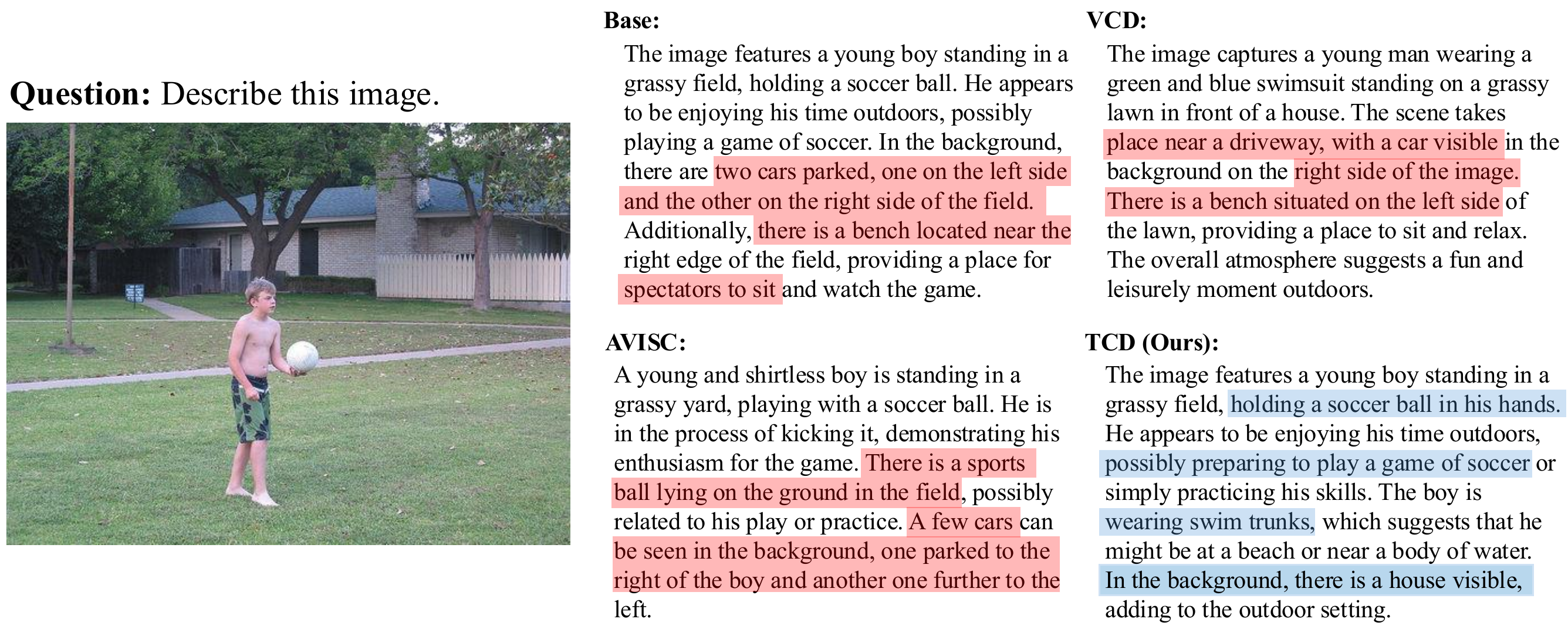}
    \vspace{-.5em}
    \caption{
        Example generated by Base, VCD~\cite{leng2023VCD}, AVISC~\cite{woo2024dontmissforesttrees} and our method, given the question: ``Describe this image.'' We randomly sample from Amber~\cite{wang2023llm} and all results are based on LLaVA-v1.5-7b model. We highlight phrases with \colorbox{red!20}{red} if they are \textit{not} well visually grounded, and with \colorbox{blue!15}{blue} if they are well visually grounded. We observe that our model successfully mitigates hallucinations compared to the other three baselines. Additional examples are provided in Appendix~\cref{fig: Qualitative_3}.
    }
    \vspace{-.25em}
    \label{fig: Qualitative_2}
\end{figure*}




\begin{figure}[tb!]
    \centering
    \includegraphics[width=.95\linewidth]{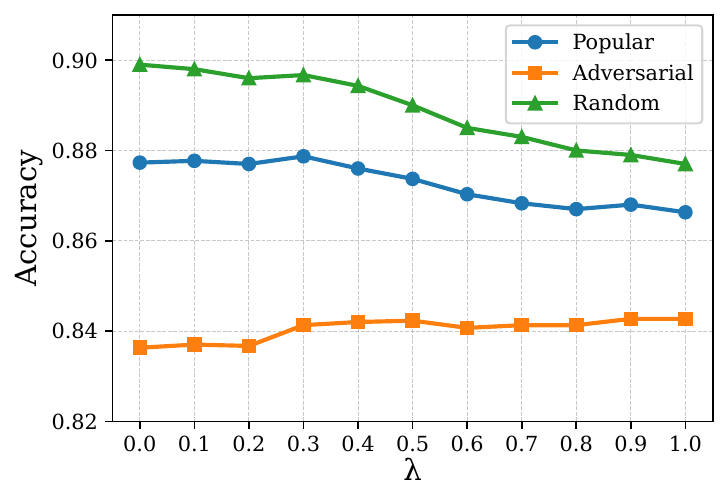}
    \vspace{-.75em}
    \caption{
    Comparison of accuracy across subsets of POPE-MSCOCO under varying $\lambda$ in the ablation setup, based on \cref{eq: lambda-ablation}. While the Random and Popular subsets show improved performance when the visual layer dominates (i.e., lower $\lambda$), the Adversarial subset benefits from a greater contributions of the amateur layer (i.e., larger $\lambda$), highlighting the distinct roles of the visual and amateur layers in mitigating different forms of hallucination.
    }
    \vspace{-.25em}
    \label{fig: lambda_ablation}
\end{figure}


\myparagraph{Visual Layer Selection Analysis.}
\cref{fig: LLaVA_Layer} shows the results of our proposed visually grounded layer search, demonstrating its robustness through consistent layer selection across the model. We also observe noticeable variation across different models, implying representational differences among LVLMs.

\myparagraph{Qualitative Analysis.}
As depicted in the \cref{fig: Qualitative_2}, our method successfully mitigated hallucinations in the original model, increase in factuality, describing number of people correctly where other methods suffer. While other models generates similar hallucinations (i.e., cars in the background), likely to be triggered from memorization of the training data, our method distinguishes the house visible in the background.

\myparagraph{Tri-layer Selection Analysis.}
\cref{tab: main_13b} shows that contrasting the visual layer (+VL) with amateur layer (+AL) consistently boosts F1, except in the adversarial split. To isolate each layer’s role, we interpolate the logits as follows:
\begin{equation}
\label{eq: lambda-ablation}
z^{(L)} - \lambda z^{(l_a)} + (1 - \lambda) z^{(l_v)},
\end{equation}
and sweep $\lambda$. \cref{fig: lambda_ablation} highlight the distinct roles played by each layer in our tri-layer decoding framework. In Random and Popular subsets, accuracy increases as $\lambda$ decreases, emphasizing the importance of the visually grounded layer $l_v$ in typical scenarios. Conversely, the Adversarial subset benefits from larger $\lambda$, as the amateur layer $l_a$ injects a complementary distribution less biased by co-occurrence patterns learned during pretraining \cite{chuang2024dola}. This helps mitigate hallucinations triggered by visually plausible yet incorrect objects. These results suggest that our tri-layer formulation effectively addresses two major sources of hallucination commonly discussed in LVLMs: (i) internal linguistic biases and (ii) weak visual grounding. The JSD-guided selection of $l_a$ helps counteract the former, especially in adversarial contexts, while the watermark guided $l_v$ enhances visual alignment in standard inputs. While we fix $\lambda$ for simplicity in our main results, the ablation findings suggest promising directions for adaptive weighting strategies based on input characteristics.

\section{Conclusion}
In this paper, we introduce Tri-layer Contrastive Decoding (TCD), a training-free framework for reducing hallucinations in Large Vision-Language Models (LVLMs). Rather than assuming the final model output always provides the best visual grounding, we propose a principled approach that embeds lightweight visual watermarks into input images and leverages targeted visual queries to probe layer-wise consistency. By combining this watermark-guided visual layer selection with contrastive decoding across mature, amateur, and visually grounded layers, TCD dynamically recalibrates the model’s reliance on vision and language, significantly improving factuality. 
\section{Limitations}
While our method demonstrates consistent improvements across multiple benchmarks and models, several limitations remain. First, our layer selection mechanism is intentionally simple and interpretable, relying on fixed, rule-based comparisons of intermediate logits. This choice benefits reproducibility and transparency, but more sophisticated or learned strategies—such as attention-based routing or score aggregation—could further enhance flexibility and robustness, especially for models with more complex encoder-decoder architectures. Additionally, extending interpretability beyond decoder layers to the visual encoder itself remains an open and promising direction.

Second, our current implementation requires multiple decoding passes to evaluate candidate layers. Although inference can be reduced to a single pass if the preferred layer is predefined or learned, developing a seamless and fully dynamic layer selection mechanism without multi-pass exploration is still an open challenge.

Third, for generation tasks, we follow AMBER’s non-LLM-based evaluation protocol to reduce subjectivity and improve reproducibility. While this is consistent with prior literature, it limits direct comparison to studies that use LLM-based scoring. Developing a more robust evaluation framework—balancing reproducibility with semantic depth, for example via ensemble metrics or human-in-the-loop evaluation—would further strengthen future studies on hallucination mitigation.

Further discussions regarding baselines and experimental settings are provided in~\cref{baseline-selection}.


\section{Ethics Statement}
All experiments are conducted using publicly available datasets (POPE, MME, AMBER), none of which contain personally identifiable or sensitive information. While our method aims to reduce object hallucinations by improving visual grounding, it does not address other potential biases—such as social, demographic, or ethical biases—that may already exist in the underlying LVLMs. In certain cases, stronger visual grounding could inadvertently reinforce existing biases.

\myparagraph{Acknowledgements} 
This work was the result of project supported by KT(Korea Telecom)-Korea University AICT R\&D Center. Also, this work was supported by IITP grant funded by the Korea government (MSIT) (IITP-2025-RS-2024-00397085, 15\%, RS-2025-02263754, Human-Centric Embodied AI Agents with Autonomous Decision-Making, 10\%, IITP-2025-RS-2020-II201819, 5\%, RS-2022-II220043, Adaptive Personality for Intelligent Agents, 10\%). We also thank Janghyun Baek and
Miso Choi for their helpful discussions and feedback.

\bibliography{main}

\appendix
\newpage
\label{sec:appendix}

\begin{table*}[tb!]
\centering
\resizebox{\textwidth}{!}{
\scriptsize
\renewcommand{\arraystretch}{1.15}
\begin{tabular}{lll}
\toprule
Dataset & Model & Template \\
\midrule
POPE / MME & LLaVA-1.5 & 
{\ttfamily <question>\textbackslash n Please answer the question using a single word or phrase.} \\
\midrule
POPE / MME & InstructBLIP & 
{\ttfamily <ImageHere> <question> Short answer:} \\
\midrule
AMBER & All & 
{\ttfamily Describe this image.} \\
\bottomrule
\end{tabular}
}
\caption{Prompt templates used for each dataset–model pair. All baselines and our method use the identical text prompt.}
\label{tab:prompt}
\end{table*}

\section{Ablation Study on Watermark Parameters}

\begin{figure}[t]
    \centering
    \includegraphics[width=\linewidth]{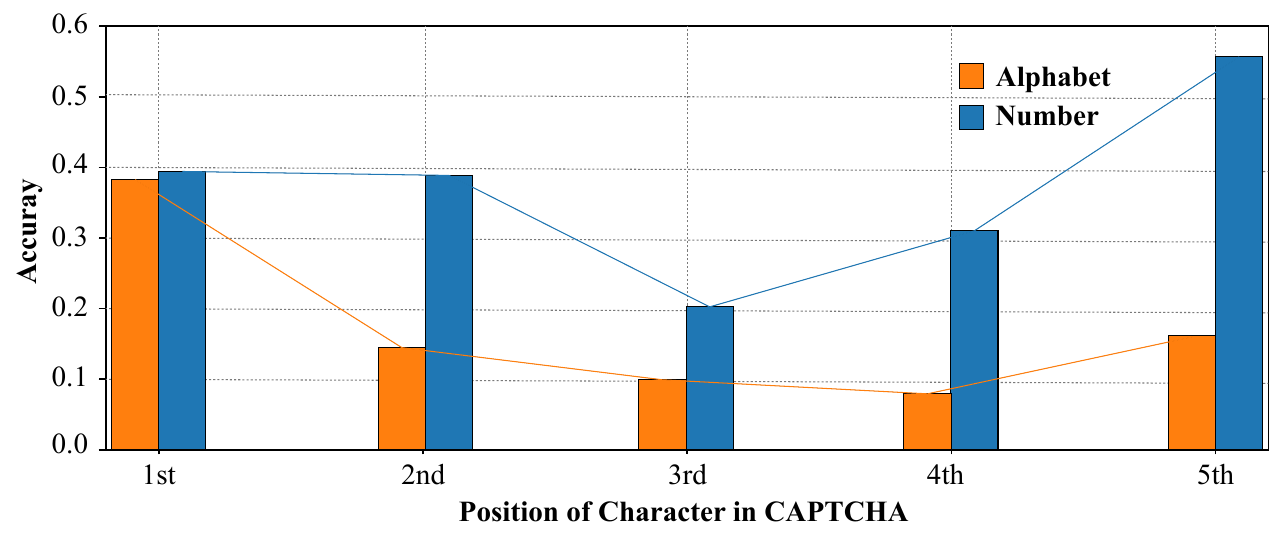}
    \caption{
        Qualitative result of CAPTCHA position. LVLM tends to answer numbers better than alphabet, last fifth character better than the other position. 
    }
    \label{fig: CAPTCHA}
\end{figure}

\myparagraph{Visual Grounding Question and CAPTCHA selection.} 
Since the key of tri-layer contrastive decoding is to select a visually grounded pivot layer with early exit token prediction method, ``a well designed question'' that judges a layer robustly is crucial. Since the LVLM utilizes the LLM, it is sensitive to both the textual and visual input queries. If we design a task that is simple, the token probability may not be meaningful to choose a pivot layer. From this perspective, we chose CAPTCHA~\cite{captcha}  as a suitable complex visual input. Together with the visual query, we conducted a simple experiment with  to fix both the image and text question. As shown in~\cref{fig: CAPTCHA}, we found that LVLM (i.e., LLaVA-1.5) tends to answer the last captcha character better. With some more finding such that LVLMs tend to have problems with recognizing numbers such as ``0'', ``9'' that may resemble the alphabet letters, we chose ``f6ww8'' as our experiment CAPTCHA. With these experiments, we fixed the question that select the visual-grounded layer as ``What is the last captcha number in the image?''.

\section{Artifacts}
\subsection{Prompt Template}
For each benchmark, we follow the official prompt template. For LLaVA-1.5, we adopt the POPE/MME instruction ending with \emph{“Please answer the question using a single word or phrase.”}, a commonly used template for short answer generation of LVLM. For InstructBLIP, we follow its native \emph{Short answer} scheme, which explicitly separates the image placeholder from the question. AMBER is designed as an open-ended description benchmark, so we keep its original single-sentence prompt. See \cref{tab:prompt} for detail.

\label{watermark implementation}
\algnewcommand\algorithmicinput{\textbf{Input:}}
\algnewcommand\algorithmicoutput{\textbf{Output:}}
\algnewcommand\algorithmiclet{\textbf{Let}}
\algnewcommand\Input{\item[\algorithmicinput]}
\algnewcommand\Output{\item[\algorithmicoutput]}
\algnewcommand\Let{\item[\algorithmiclet]}

\begin{algorithm}[tb!]
    \caption{~Embedding Visible Identifier (Watermarking)}
    \label{alg: watermark}
    \begin{algorithmic}[1]
        \Input
            original image $\mathcal{I_{\text{o}}}$, watermark image $\mathcal{I_{\text{w}}}$, image dimensions $(x_\text{o},\,y_\text{o})$, $(x_\text{w},\,y_\text{w})$, and opacity $\alpha$
        \vspace{.25em}
        
        \Let $(0, 0)$ be the top-left pixel of $\mathcal{I}_\text{o}$, and $\mathcal{C}_\text{w}=(c_{\text{w}}^{(x)},\,c_{\text{w}}^{(y)})$ be the center pixel of $\mathcal{I}_\text{w}$,
        \State $\mathcal{P}_\text{o} \gets (0.9x_\text{o},\,0.9y_\text{o})$ \Comment{bottom-right anchor pixel}
        \State $\mathcal{C}_\text{w} \gets \mathcal{P}_\text{o}$ \Comment{overlapping watermark}
        \vspace{.25em}
        \While{$\mathcal{C}_{\text{w}} + (x_\text{w}/2,\,y_\text{w}/2) > (x_\text{o},\,y_\text{o})$}
            \If{$c_{\text{w}}^{(x)} + x_\text{w}/2 > x_\text{o}$} \Comment{resize width}
                \State $x_\text{w} \gets \min(x_\text{w}/2,\, x_\text{o}-x_\text{w})$ 
            \EndIf
            \If{$c_{\text{w}}^{(y)} + y_\text{w}/2 > y_\text{o}$} \Comment{resize height}
                \State $y_\text{w} \gets \min(y_\text{w}/2,\, y_\text{o}-y_\text{w})$ 
            \EndIf
        \EndWhile

        \State $\mathcal{I} \gets \mathcal{I}_\text{o} + \alpha\mathcal{I_{\text{w}}}$ \Comment{watermark integration}
        \vspace{.5em}

        \Output watermark-embedded image $\mathcal{I}$
    \end{algorithmic}
\end{algorithm}

\begin{figure}[tb!]
    \centering
    \includegraphics[width=\linewidth]{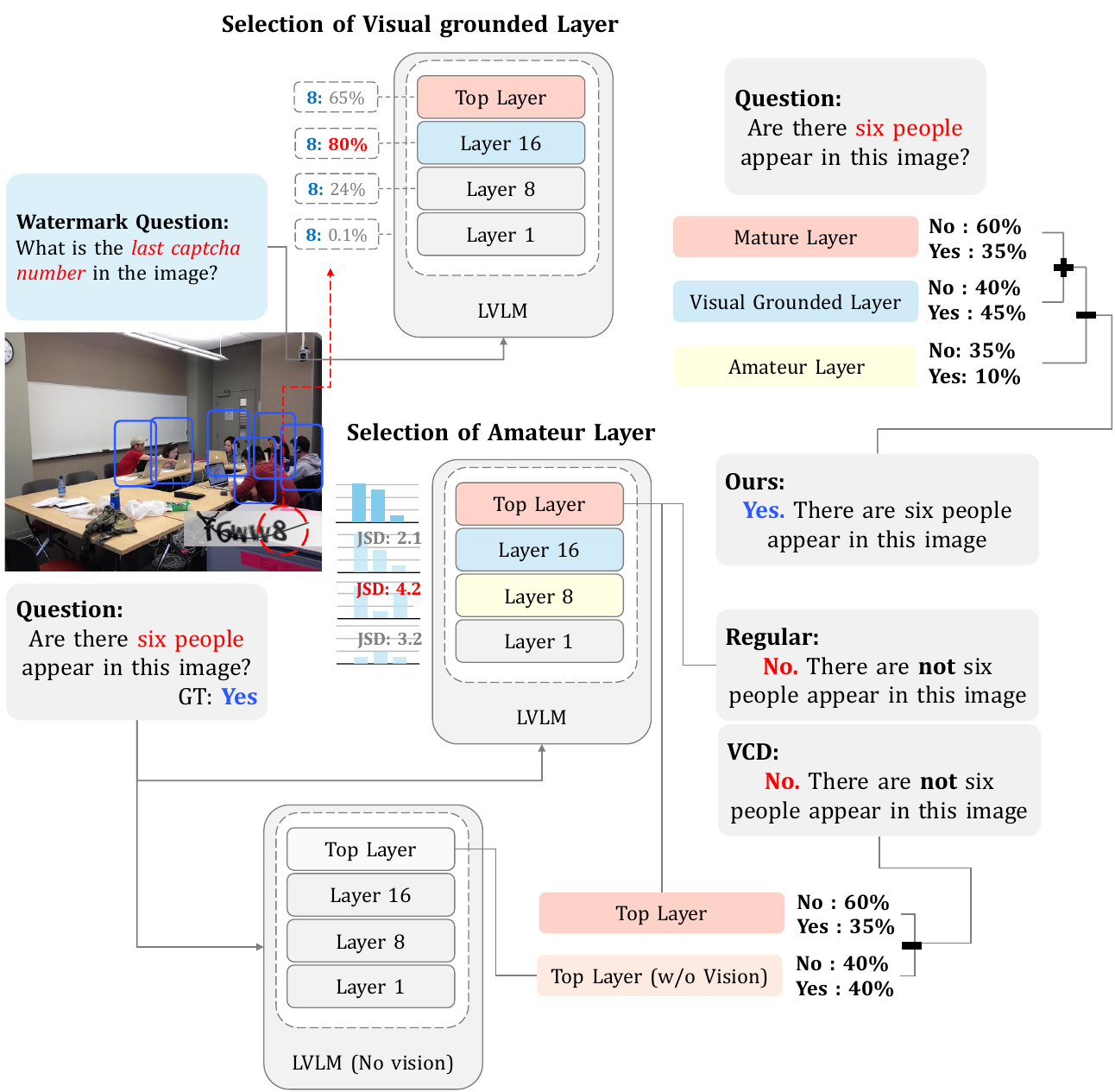}
    \caption{
        Examples of our tri-layer contrastive decoding approach on a sample from MME benchmark. We observe that our model outperforms the other alternatives, i.e., VCD~\cite{leng2023VCD}, successfully mitigating hallucinations that LVLMs suffer. Note that an original image without watermark is used for all methods.
    }
    \label{fig: Tri-layer-over}
\end{figure}

\begin{table}[h]
    \footnotesize
    \centering
    \begin{tabular*}{\linewidth}{@{\extracolsep{\fill}}lccc}
        \toprule
        \multirow{2}{*}{\centering Model} & \multicolumn{3}{c}{\textbf{Perception Score ($\uparrow$)}} \\
        \cmidrule{2-4}
        & Regular & VCD & \textbf{Ours} \\
        \midrule
        LLaVA1.5 & 1277.6 & 1338.2 & \textbf{1500.4} \textcolor{red}{(+162.2)} \\
        InstructBLIP & 1050.9 & 1202.2 & \textbf{1240.73} \textcolor{red}{(+38.53)} \\
        \bottomrule
    \end{tabular*}
    \caption{
        Evaluation of hallucination using various models and decoding methods on the coarse-grained perception subset of MME~\cite{fu2024MME} benchmark.
        The best performances are \textbf{bolded}.
    }
    \label{tab:sub_mme}
\end{table}

\section{Additional Implementation Details}
\label{app: C}
\subsection{Hardware and Software Environment}
All experiments with LLaVA v1.5 were conducted using PyTorch~2.1.2, CUDA~12.1, while InstructBLIP experiments relied on PyTorch~2.0.1, CUDA~11.7. The two configurations reflect the official code bases: LLaVA~\cite{liu2024LLaVa1.5} and OPERA (the reference implementation of InstructBLIP)~\cite{Huang_2024_CVPR}. Unless otherwise noted, inference and evaluation were run on a single NVIDIA~RTX~A6000 (48\,GB). Experiments with DeepSeek-VL2-Tiny were executed on an NVIDIA~H100~NVL.

\subsection{Hyper-parameter Configuration}
Table~\ref{tab:hyperparams} lists the hyper-parameters used for every dataset–scenario–model combination. For each dataset we fix a single configuration and reuse it across the Random, Popular, and Adversarial splits to ensure a fair comparison. Although tuning the parameters per sample or subset can yield higher scores, our objective here is to show that \textbf{visually grounded} tri-layer selection is feasible; achieving optimal performance is left to future work.

\begin{table}[h]
\centering
\resizebox{\linewidth}{!}{
\begin{tabular}{lcccc}
\toprule
Model & Dataset (Split) & $\lambda$ & Gain Search & Candidate $k$ \\
\midrule
\multirow{11}{*}{LLaVA-1.5}
& MSCOCO (Random)      & 1.0 & change & 20 \\
& MSCOCO (Popular)     & 1.0 & change & 20 \\
& MSCOCO (Adversarial) & 1.0 & change & 20 \\
& AOK-VQA (Random)     & 0.5 & log    & 20 \\
& AOK-VQA (Popular)    & 0.5 & log    & 20 \\
& AOK-VQA (Adversarial)& 0.5 & log    & 20 \\
& GQA (Random)         & 0.1 & log    & 20 \\
& GQA (Popular)        & 0.1 & log    & 20 \\
& GQA (Adversarial)    & 0.1 & log    & 20 \\
& MME (–)              & 0.5 & change & 20 \\
& AMBER (–)            & 0.5 & log    & 20 \\
\midrule
\multirow{11}{*}{InstructBLIP}
& MSCOCO (Random)      & 0.3 & change & 20 \\
& MSCOCO (Popular)     & 0.3 & change & 20 \\
& MSCOCO (Adversarial) & 0.3 & change & 20 \\
& AOK-VQA (Random)     & 0.3 & change & 20 \\
& AOK-VQA (Popular)    & 0.3 & change & 20 \\
& AOK-VQA (Adversarial)& 0.3 & change & 20 \\
& GQA (Random)         & 0.3 & change & 20 \\
& GQA (Popular)        & 0.3 & change & 20 \\
& GQA (Adversarial)    & 0.3 & change & 20 \\
& MME (–)              & 1.0 & log    & 10 \\
& AMBER (–)            & 0.5 & log    & 20 \\
\bottomrule
\end{tabular}}
\caption{Hyper-parameters for all dataset–scenario combinations. A single configuration per dataset is reused across splits to enable consistent comparison.}
\label{tab:hyperparams}
\end{table}

\subsection{Implementation on stronger backbone}
We additionally evaluate our method on the AMBER benchmark using \textsc{DeepSeek}-VL2-Tiny, a Mixture-of-Experts model with a substantially stronger backbone than Vicuna-7B despite its smaller parameter count (3.37 B). For the VCD baseline~\cite{leng2023VCD}, we follow the authors’ recommendations and sweep \(\alpha = 1.0\) while varying \(\beta \in [0.2,\,0.5]\); we report the best AMBER score obtained. For TCD, we treat the last eight decoder layers (of twelve) as candidates and select layer 4 as the visually grounded pivot, based on a preliminary sweep with a small watermarking subset.

\begin{table}[tb!]
    \footnotesize
    \centering
    \begin{tabular}{lcc}
        \toprule
        Method            & Latency (s) ($\downarrow$)         & Throughput (tk/s) ($\uparrow$)    \\
        \midrule
        LLaVA-1.5-7B     & 0.17 $\pm$ 0.06      & 32.89 $\pm$ 3.68          \\
         + VCD               & 0.56 $\pm$ 0.03      & 17.97 $\pm$ 0.83          \\
         + AVISC             & 0.28 $\pm$ 0.07      & 15.93 $\pm$ 1.45          \\
         + VCD (Ours)        & 0.38 $\pm$ 0.01      & 26.88 $\pm$ 0.58          \\
        \bottomrule
    \end{tabular}
    \caption{ Comparison with the baseline Contrastive Decoding methods for the Latency and Throughput (tokens/s).
    }
\end{table}

\section{Latency}
We report decoding latency (seconds) and throughput (tokens per second, t/s; mean~$\pm$~standard deviation) on the AMBER generation task. Eleven samples were drawn at random, and the first sample in each run
was discarded to avoid warm-up bias. All methods were executed with their \emph{official} implementations on a single NVIDIA~H100 GPU, using a batch size of one and a maximum generation length of ten tokens. Our method evaluates $k=20$ candidate layers per decoding step. 

\section{Discussion of Baseline Selection}
\label{baseline-selection}
As discussed in \cref{subsec: quant_analysis}, we selected VCD, M3ID, and AVISC as our primary training-free contrastive decoding baselines, and included ICD, EOS~\cite{yue-etal-2024-less}, HA-DPO~\cite{zhao2024hallucinationsenhancinglvlmshallucinationaware}, HALVA~\cite{sarkar2025mitigating}, and Octopus as reference methods that require additional training or external modules. Nonetheless, there exist other notable variations in decoding-based approaches for mitigating hallucinations in LVLMs. For example, PAI~\cite{liu2024paying} proposes a method similar to VCD, introducing visual perturbations to strengthen visual input, while ConVis~\cite{Park_Lee_Choe_Chang_2025} leverages SDXL, a text-to-image model, to further ground LVLMs using generated images.

Given the diversity of possible experimental setups—such as model choices (e.g., LLaVA-1.5, InstructBLIP, QwenVL~\cite{bai2023qwenvlversatilevisionlanguagemodel}, MiniGPT~\cite{zhu2023minigpt4enhancingvisionlanguageunderstanding}, and Shikra~\cite{chen2023shikraunleashingmultimodalllms}), benchmarks and evaluation metrics (e.g., POPE-MSCOCO, POPE-OKVQA, POPE-GQA, MME-Perception (example on \cref{fig: Tri-layer-over}), MME-Cognition, AMBER, CHAIR, MMVP~\cite{zhong2023mmvpmotionmatrixbasedvideoprediction}, and MMbench~\cite{liu2024mmbenchmultimodalmodelallaround}), we aimed to align our experimental design with the conventions established by recent works such as Octopus\cite{suo2025octopusalleviatinghallucinationdynamic} and AVISC\cite{woo2024dontmissforesttrees}. For instance, although reporting results for each POPE subset independently could highlight the strengths of our method, we chose to aggregate all POPE subsets into a single evaluation to provide a fair and comprehensive comparison, as recommended by recent literature.

\begin{figure*}[tb!]
    \centering
    \includegraphics[width=\linewidth]{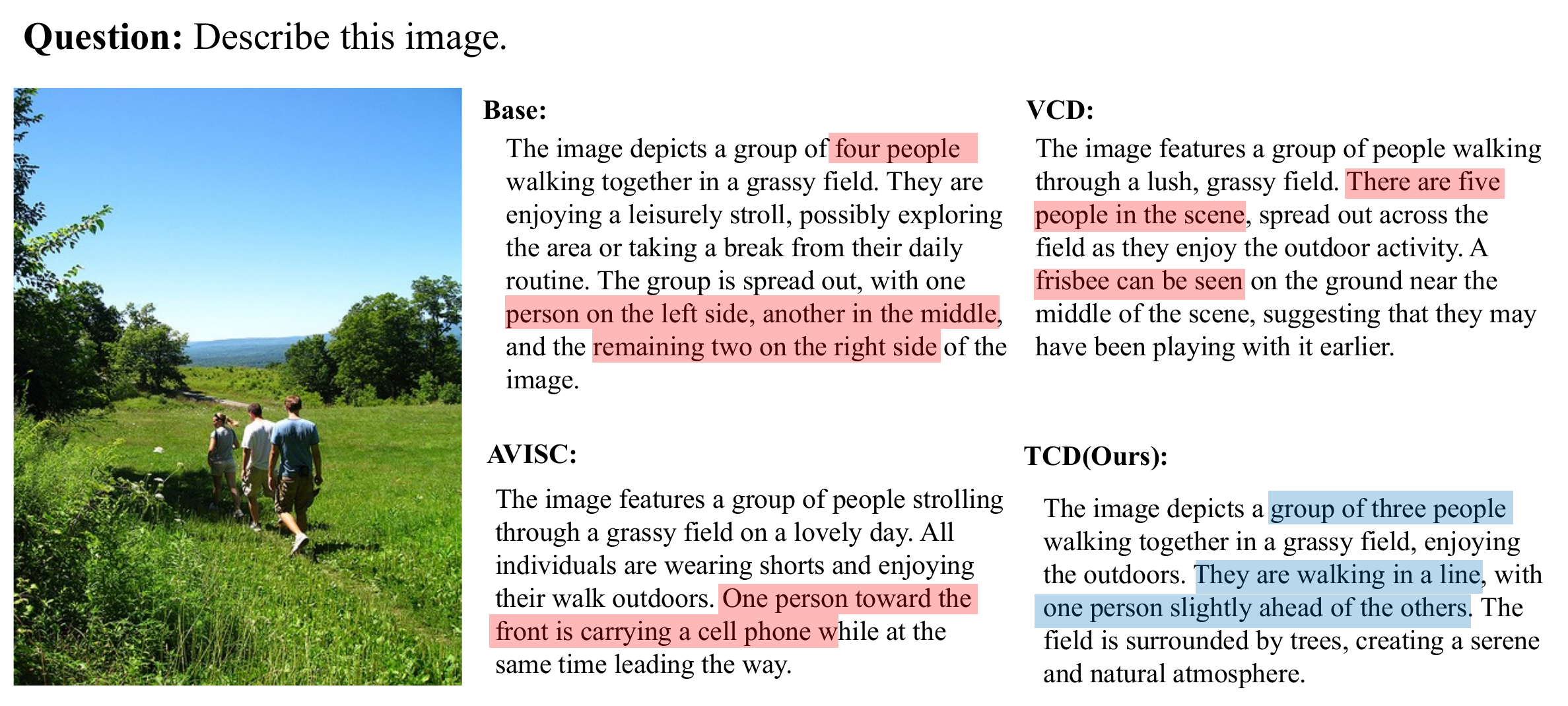}
    \includegraphics[width=\linewidth]{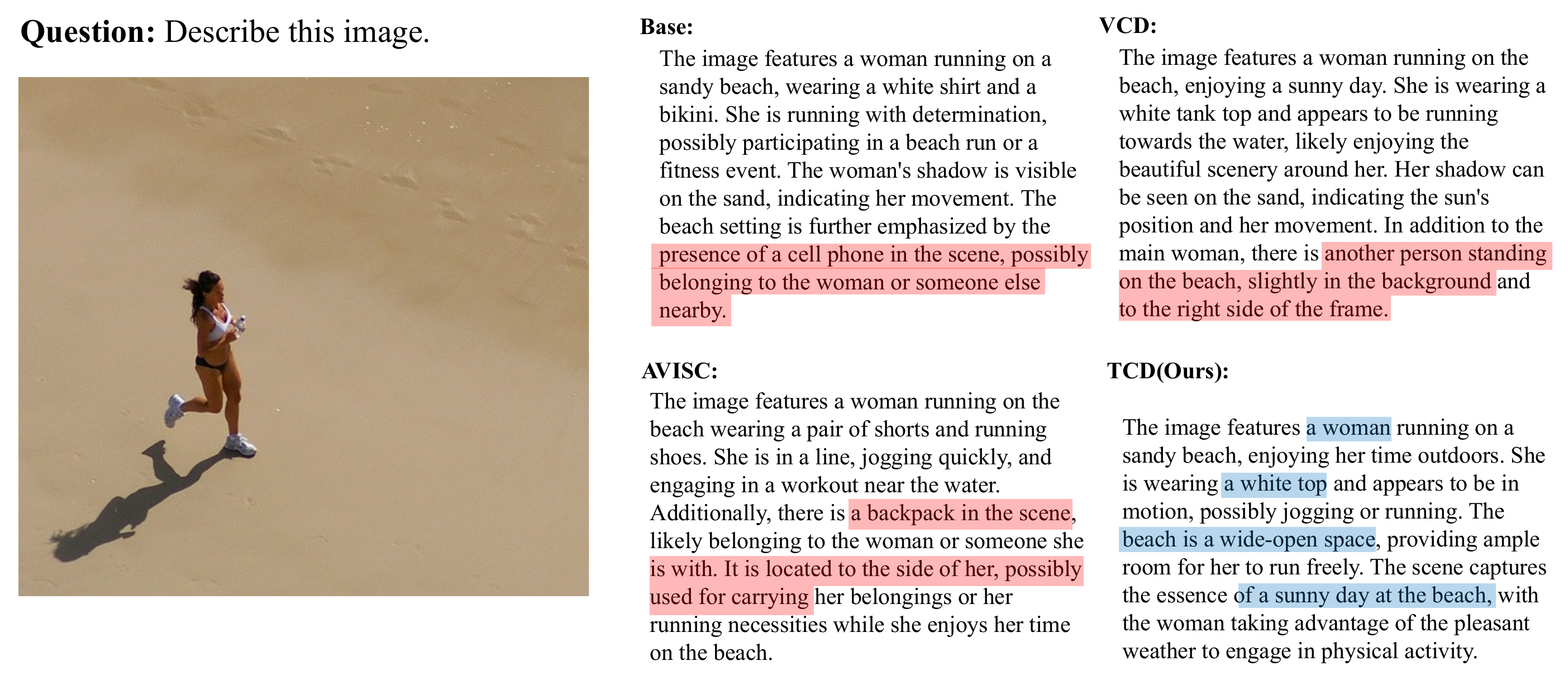}
    \includegraphics[width=\linewidth]{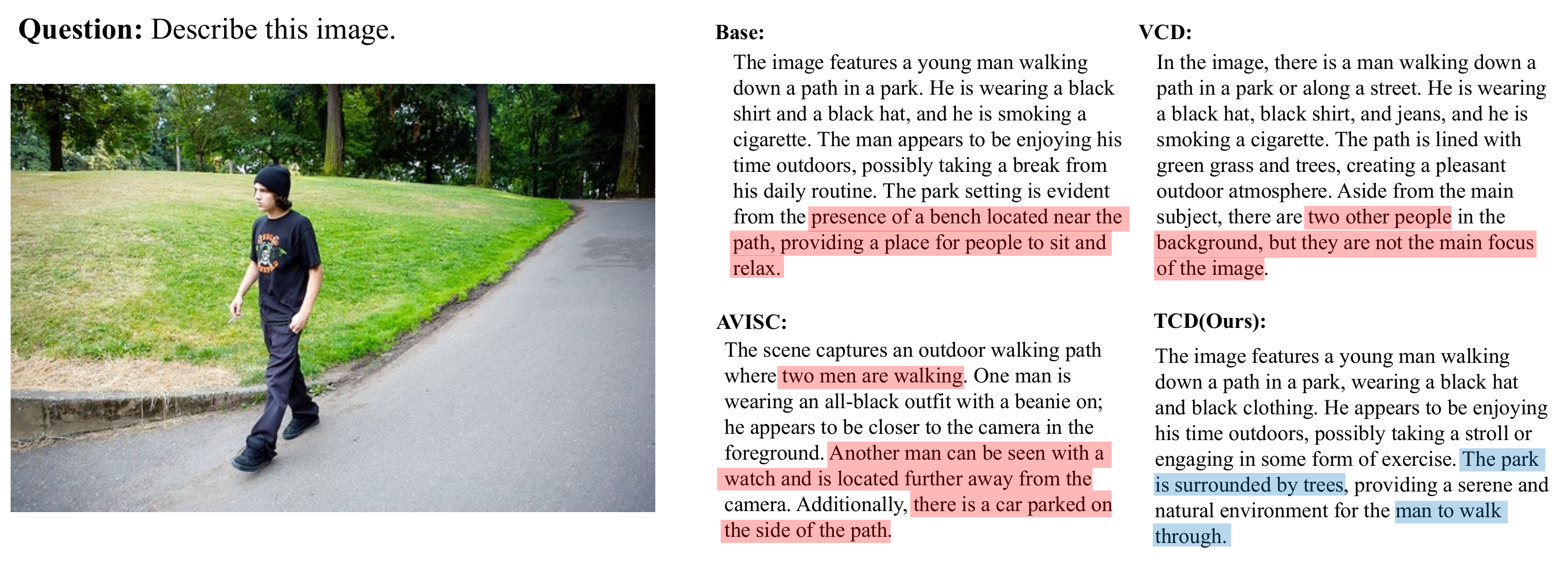}
    \caption{
        More examples generated by Base, VCD~\cite{leng2023VCD}, AVISC~\cite{woo2024dontmissforesttrees} and ours given a question, such as ``Describe this image.'' We extract random samples from Amber~\cite{wang2023llm} and all results are based on LLaVA v1.5 7b. We highlight phrases with \textcolor{red}{red} if it is \textit{not} well visually grounded, and with \textcolor{blue}{blue} if it is well visually grounded. We observe that our model successfully mitigates hallucinations compared to other three baselines. 
    }
    \label{fig: Qualitative_3}
\end{figure*}

\section{Additional Experimental Results}

This section provides supplementary results, including a detailed ablation study on watermark properties and an evaluation on a free-form generation benchmark.

\subsection{Ablation Study on Watermark Design}
\label{sec:appendix_ablation}

To assess the sensitivity of our method to watermark design, we conducted an ablation study on the POPE-MSCOCO benchmark with LLaVA-1.5-7B. We independently varied the watermark's opacity, position, and size. The results, summarized in Tables \ref{tab:ablation_combined}, confirm that a moderately sized ($1.0\times$), semi-transparent (0.8 opacity) watermark placed in the bottom-right corner is optimal. This configuration maximizes factual accuracy (Accuracy and F1 Score) while ensuring that the visually grounded layer is consistently identified within the mid-depth blocks of the model, indicating a stable and reliable grounding process.

\subsection{Evaluation on Free-form Generation}
\label{sec:appendix_freeform}

To address concerns about the potential impact of our method on the MLLM's general capabilities, we conducted an additional evaluation on the LLaVA-wild benchmark using GPT-4 evaluation. As shown in Table \ref{tab:llava_wild_eval}, our method (TCD) not only preserves but enhances the model's performance. Notably, there is a significant improvement in the "Detail description" category, suggesting that by improving factual grounding, TCD enables the model to generate more precise and detailed descriptions without compromising its conversational or reasoning abilities.

\begin{table*}[t]
    \centering
    \setlength{\tabcolsep}{6pt} 
    \renewcommand{\arraystretch}{1.2} 
    \begin{tabular}{@{}lcccc@{}}
    \toprule
    \textbf{Model} & \textbf{Conversation} & \textbf{Detail Description} & \textbf{Complex Reasoning} & \textbf{Overall Score} \\
    \midrule
    LLaVA-v1.5-7B & 60.3 & 39.0 & 71.7 & 59.9 \\
    \rowcolor{gray!10}
    \textbf{+ TCD} & \textbf{61.2 (+1.1)} & \textbf{45.4 (+6.4)} & \textbf{72.1 (+0.4)} & \textbf{62.2 (+2.3)} \\
    \bottomrule
    \end{tabular}
    \vspace{0.3em}
    \caption{Evaluation on the LLaVA-wild benchmark, with scores obtained via LLM-eval (GPT-4o). TCD improves performance across all categories compared with the LLaVA-v1.5-7B baseline. A particularly strong gain is observed in generating detailed descriptions (+6.4), demonstrating TCD's effectiveness in enhancing the model's visual descriptive capabilities.}
    \label{tab:llava_wild_eval}
\end{table*}

\begin{table*}[t]
    \centering
    \setlength{\tabcolsep}{8pt} 
    \renewcommand{\arraystretch}{1.15} 
    \resizebox{\linewidth}{!}{%
    \begin{tabular}{@{}llcccccccccc@{}}
    \toprule
    \multirow{2}{*}{\textbf{Ablation}} & \multirow{2}{*}{\textbf{Setting}} & \multirow{2}{*}{\textbf{Accuracy}} & \multirow{2}{*}{\textbf{F1 Score}} & \multicolumn{8}{c}{\textbf{Layer Selection Frequency}} \\
    \cmidrule(lr){5-12}
    & & & & None$^*$ & L26 & L27 & L28 & L29 & L30 & L31 & L32 \\
    \midrule
    
    \multirow{3}{*}{\textbf{Size}} 
    & $0.5\times$ & 0.8482 & 0.8340 & -- & 7 & 31 & 20 & 64 & 92 & 0 & 43 \\
    & $2.0\times$ & 0.8685 & 0.8644 & -- & 10 & 2 & -- & 37 & 1 & -- & 261 \\
    \rowcolor{gray!10}
    & \textbf{$1.0\times$} & \textbf{0.8700} & \textbf{0.8665} & -- & \textbf{3} & \textbf{2} & \textbf{2} & \textbf{168} & -- & \textbf{82} & -- \\
    \midrule
    
    \multirow{3}{*}{\textbf{Position}} 
    & Center & 0.8515 & 0.8380 & -- & 9 & -- & 23 & 82 & 104 & 14 & 25 \\
    & Random & 0.8500 & 0.8348 & -- & 10 & -- & 18 & 37 & 117 & 18 & 21 \\
    \rowcolor{gray!10}
    & \textbf{Bottom Right} & \textbf{0.8700} & \textbf{0.8665} & -- & \textbf{3} & -- & \textbf{22} & \textbf{168} & -- & \textbf{82} & -- \\
    \midrule
    
    \multirow{5}{*}{\textbf{Opacity}} 
    & 0.0 & 0.8339 & 0.8065 & 1 & -- & -- & 2 & -- & 4 & 39 & 58 \\
    & 0.2 & 0.8351 & 0.8081 & 2 & -- & -- & 1 & -- & 4 & 37 & 60 \\
    & 0.4 & 0.8329 & 0.8060 & -- & 2 & -- & -- & 3 & 8 & 43 & 71 \\
    & 0.6 & 0.8638 & 0.8563 & -- & 3 & -- & 43 & 27 & 16 & 69 & 52 \\
    \rowcolor{gray!10}
    & \textbf{0.8} & \textbf{0.8700} & \textbf{0.8665} & -- & \textbf{3} & -- & \textbf{22} & \textbf{168} & -- & \textbf{82} & -- \\
    \bottomrule
    \end{tabular}}
    \vspace{0.3em}
    \caption{
        Comprehensive ablation studies on watermark properties: size, position, and opacity. 
        Performance is evaluated using LLaVA-1.5v-7b model with POPE MSCOCO dataset, alongside the frequency of layer selection for visual grounding.
        Our final chosen settings are as follows : (\textbf{size: $1.0\times$}, \textbf{position: Bottom Right}, \textbf{opacity: 0.8}) .
        $^*$None: Indicates cases where no layer produced the correct answer token.
    }
    \label{tab:ablation_combined}
\end{table*}

\begin{table*}[t] 
    \centering
    \setlength{\tabcolsep}{15pt}
    \renewcommand{\arraystretch}{0.9}
    \resizebox{\linewidth}{!}{%
    \begin{tabular}{@{}lcccccccccccc@{}}
        \toprule
        \multirow{2}{*}{\textbf{Method}} &
        \multicolumn{2}{c}{Random} &
        \multicolumn{2}{c}{Popular} &
        \multicolumn{2}{c}{Adversarial} &
        \multicolumn{2}{c}{ALL} \\
        \cmidrule(lr){2-3} \cmidrule(lr){4-5} \cmidrule(lr){6-7} \cmidrule(lr){8-9}
        & Acc. & F1 & Acc. & F1 & Acc. & F1 & Acc. & F1 \\
        \midrule
        LLaVA-1.5-7B             & 83.77          & 81.94          & 82.57           & 80.86          & 79.77            & 78.47            & 82.04          & 80.42          \\
        \quad +ICD               & \underline{87.51} & 83.28          & 83.15        & 83.91          & 79.13            & 80.41 & 83.26          & 82.53 \\
        \quad +ConVis            & 84.70          & –              & 83.20           & –              & 81.10            & –                & 83.00          & –              \\
        \quad +OPERA             & 84.40          & –              & 83.40           & –              & 81.20            & –                & 83.00          & –              \\
        \quad +VCD               & 85.43          & 83.99          & 83.17           & 81.94          & 80.27            & 79.49            & 82.96          & 81.81          \\
        \quad +M3ID$^\dag$       & 86.13          & 81.85          & 82.07           & 80.77          & 79.50            & 78.15            & 82.57          & 80.26          \\
        \quad +AVISC             & 84.67          & 82.21          & 83.67           & 81.27          & 81.83 & 79.55            & 83.39  & 81.01          \\
        \quad +Octopus           & \underline{87.51} & \underline{85.40} & \underline{85.20} & \underline{84.19} & \underline{82.22} & \underline{81.44} & \underline{85.79} & \underline{83.44} \\
        \rowcolor{gray!10}
        \quad TCD (Ours)         & \textbf{89.50} & \textbf{88.89} & \textbf{87.60}  & \textbf{87.14} & \textbf{83.90}   & \textbf{83.92}   & \textbf{87.00} & \textbf{86.65} \\
        \midrule
        InstructBLIP             & 81.53          & 81.19          & 78.47           & 78.75          & 77.43            & 78.00            & 79.14          & 79.31          \\
        \quad +ICD               & 84.36          & 83.82          & 77.88           & 78.70          & 75.17            & 77.23            & 79.14          & 79.92          \\
        \quad +OPERA             & 84.57          & 83.74          & 78.24           & 79.15          & 74.59            & 76.33            & 79.13          & 79.74          \\
        \quad +VCD               & 82.03          & 81.56          & 79.13           & 79.20          & 77.23            & 77.72            & 79.46          & 79.49          \\
        \quad +M3ID$^\dag$       & 82.33          & 81.53          & 80.90           & 80.42          & 78.53            & 78.49            & 80.59          & 80.15          \\
        \quad +AVISC             & 86.03          & 84.41          & \underline{84.27} & \underline{82.77} & \underline{81.83} & 80.67            & 84.04          & 82.62          \\
        \quad +Octopus           & \underline{86.63} & \underline{85.30} & \textbf{84.90}  & \textbf{83.55} & \textbf{82.83}   & \textbf{81.43}   & \textbf{84.79} & \underline{83.43} \\
        \rowcolor{gray!10}
        \quad TCD (Ours)         & \textbf{88.40} & \textbf{87.63} & 82.77           & 82.67          & 81.13            & \underline{81.33} & \underline{84.10} & \textbf{83.88} \\
        \bottomrule
    \end{tabular}}
    \caption{
    Comparison with the state-of-the-art methods for the discriminative tasks on the POPE\_MSCOCO dataset.}
    \label{tab: app_pope_mscoco}
\end{table*}

\begin{table*}[t]
    \centering
    \setlength{\tabcolsep}{15pt}
    \renewcommand{\arraystretch}{0.9}
    \resizebox{\linewidth}{!}{%
    \begin{tabular}{@{}lcccccccc@{}}
    \toprule
    \multirow{2}{*}{\textbf{Method}} &
    \multicolumn{2}{c}{Random} &
    \multicolumn{2}{c}{Popular} &
    \multicolumn{2}{c}{Adversarial} &
    \multicolumn{2}{c}{ALL (Avg.)} \\
    \cmidrule(lr){2-3} \cmidrule(lr){4-5} \cmidrule(lr){6-7} \cmidrule(lr){8-9}
    & Acc. & F1 & Acc. & F1 & Acc. & F1 & Acc. & F1 \\
    \midrule
    LLaVA-1.5-7B & 82.73 & 84.26 & 76.10 & 79.34 & 67.90 & 74.09 & 75.58 & 79.23 \\
    \quad +ICD & - & - & - & - & - & - & & \\
    \quad +OPERA & - & - & - & - & - & - & & \\
    \quad +VCD & 81.30 & 83.23 & 75.43 & 79.26 & 67.43 & 74.11 & 74.72 & 78.87 \\
    \quad +M3ID$^\dag$ & 83.57 & 85.09 & 76.80 & 80.06 & 68.10 & 74.58 & 76.16 & 79.91 \\
    \quad +AVISC & \underline{84.60} & \underline{85.88} & \underline{78.83} & \underline{81.63} & \underline{68.97} & \underline{75.11} & \underline{77.47} & \underline{80.87} \\
    \quad +Octopus & - & - & - & - & - & - & & \\
    \rowcolor{gray!10}\quad TCD (Ours) & \textbf{91.23} & \textbf{91.12} & \textbf{87.57} & \textbf{87.86} & \textbf{80.57} & \textbf{82.24} & \textbf{86.46} & \textbf{87.07} \\
    \midrule
    InstructBLIP & 81.00 & 82.06 & 75.00 & 77.69 & 68.80 & 73.84 & 74.93 & 77.86 \\
    \quad +ICD & - & - & - & - & - & - & & \\
    \quad +OPERA & - & - & - & - & - & - & & \\
    \quad +VCD & 81.73 & 82.66 & 75.33 & 77.92 & 69.70 & 74.27 & 75.59 & 78.28 \\
    \quad +M3ID$^\dag$ & 82.33 & 83.66 & 75.60 & 78.36 & 69.57 & 74.39 & 75.83 & 78.80 \\
    \quad +AVISC & \textbf{88.47} & \textbf{88.59} & \underline{81.77} & \underline{82.98} & \underline{72.53} & \underline{76.28} & \underline{80.92} & \underline{82.62} \\
    \quad +Octopus & - & - & - & - & - & - & & \\
    \rowcolor{gray!10}\quad TCD (Ours) & \underline{88.00} & \underline{88.36} & \textbf{84.03} & \textbf{85.08} & \textbf{76.60} & \textbf{79.56} & \textbf{82.88} & \textbf{84.33} \\
    \bottomrule
    \end{tabular}}
    \caption{Comparison with the state-of-the-art methods for the discriminative tasks on the A-OKVQA dataset.}
    \label{tab: app_pope_okvqa}
\end{table*}

\begin{table*}[t]
    \centering
    \setlength{\tabcolsep}{15pt}
    \renewcommand{\arraystretch}{0.9}
    \resizebox{\linewidth}{!}{%
    \begin{tabular}{@{}lcccccccccccc@{}}
    \toprule
    \multirow{2}{*}{\textbf{Method}} &
    \multicolumn{2}{c}{Random} &
    \multicolumn{2}{c}{Popular} &
    \multicolumn{2}{c}{Adversarial} &
    \multicolumn{2}{c}{ALL (Avg.)} \\
    \cmidrule(lr){2-3} \cmidrule(lr){4-5} \cmidrule(lr){6-7} \cmidrule(lr){8-9}
    & Acc. & F1 & Acc. & F1 & Acc. & F1 & Acc. & F1 \\
    \midrule
    LLaVA-1.5-7B            & 82.40          & 83.99          & 72.03           & 76.84          & 68.73            & 74.92            & 74.39          & 78.58          \\
    \quad +ICD              & –              & –              & –               & –              & –                & –                & –              & –              \\
    \quad +OPERA            & –              & –              & –               & –              & –                & –                & –              & –              \\
    \quad +VCD              & 82.27          & 84.22          & 71.77           & 77.05          & 68.27            & 74.84            & 74.10          & 78.70          \\
    \quad +M3ID$^\dag$      & 82.83          & 84.62          & 72.83           & 77.58          & 68.13            & 74.78            & 74.60          & 78.99          \\
    \quad +AVISC            & \underline{85.00} & \underline{86.45} & \underline{74.80} & \underline{79.17} & \underline{69.20} & \underline{75.58} & \underline{76.33} & \underline{80.40} \\
    \quad +Octopus          & –              & –              & –               & –              & –                & –                & –              & –              \\
    \rowcolor{gray!10}
    \quad TCD (Ours)        & \textbf{88.90} & \textbf{88.43} & \textbf{85.57}  & \textbf{85.46} & \textbf{81.93}   & \textbf{82.44}   & \textbf{85.47} & \textbf{85.44} \\
    \midrule
    InstructBLIP            & 80.00          & 81.02          & 73.53           & 76.49          & 68.00            & 72.59            & 73.84          & 76.70          \\
    \quad +ICD              & –              & –              & –               & –              & –                & –                & –              & –              \\
    \quad +OPERA            & –              & –              & –               & –              & –                & –                & –              & –              \\
    \quad +VCD              & 81.73          & 82.45          & 74.10           & 76.87          & 70.27            & 74.29            & 75.36          & 77.87          \\
    \quad +M3ID$^\dag$      & 80.57          & 81.85          & 74.57           & 77.53          & 68.90            & 73.47            & 74.68          & 77.62          \\
    \quad +AVISC            & \underline{86.47} & \underline{86.57} & \underline{78.00} & \underline{79.84} & \underline{73.07} & \underline{76.54} & \underline{79.85} & \underline{80.98} \\
    \quad +Octopus          & –              & –              & –               & –              & –                & –                & –              & –              \\
    \rowcolor{gray!10}
    \quad TCD (Ours)        & \textbf{86.57} & \textbf{86.79} & \textbf{80.17}  & \textbf{81.65} & \textbf{76.13}   & \textbf{78.72}   & \textbf{80.96} & \textbf{82.39} \\
    \bottomrule
    \end{tabular}}
    \caption{Comparison with the state-of-the-art methods for the discriminative tasks on the GQA dataset.}
    \label{tab: app_pope_gqa}
\end{table*}

\end{document}